\documentclass[runningheads]{llncs}

\usepackage{eccv}

\usepackage{eccvabbrv}

\usepackage{graphicx}
\usepackage{booktabs}

\usepackage[accsupp]{axessibility}  

\definecolor{junglegreen}{rgb}{0.113, 0.639, 0.5}

\definecolor{spaceblue}{rgb}{0.2,0.2,0.5}

\newcommand{\ea}[1]{{\color{black}{#1}}}

\definecolor{col1}{rgb}{0.10,0.90,0.60}

\definecolor{col2}{rgb}{0.10,0.10,0.60}

\newcommand{\acronymtitle}{SOMA}

\usepackage{booktabs}
\usepackage{pifont}
\usepackage{graphicx}
\usepackage{multirow}
\usepackage{xcolor}
\usepackage{svg}
\usepackage[dvipsnames]{xcolor}
\usepackage{wrapfig}
\usepackage[shrink=50, step=2]{microtype}

\newcommand{\cmark}{\textcolor{PineGreen}{\ding{51}}}%
\newcommand{\mmark}{\textcolor{Dandelion}{\ding{51}}}%
\newcommand{\xmark}{\textcolor{BrickRed}{\ding{55}}}%

\usepackage{hyperref}

\usepackage{orcidlink}

\begin{document}

\title{SOMA: From Surface Observations to Muscle Anatomy} 

\titlerunning{SOMA}


\author{Eduardo Alvarado\inst{1}\orcidlink{0000-0003-3395-5674}\and
Emily Kim\inst{1}\and
Gerrit Nolte\inst{2}\orcidlink{0000-0002-5080-1039}\and
Friedemann Runte\inst{2}\orcidlink{0009-0000-1963-8707}\and
Mario Botsch\inst{2}\orcidlink{0000-0001-9954-120X}\and
Marc Habermann\inst{1}\orcidlink{0000-0003-3899-7515}\and
Christian Theobalt\inst{1}\orcidlink{0000-0001-6104-6625}}

\authorrunning{E.~Alvarado et al.}


\institute{
Max Planck Institute for Informatics, Saarland Informatics Campus \\
\and
TU Dortmund University \\
[4pt]
\url{https://vcai.mpi-inf.mpg.de/projects/SOMA/}
}

\maketitle

\vspace{-7mm}
\begin{figure}
\includegraphics[width=1\textwidth]{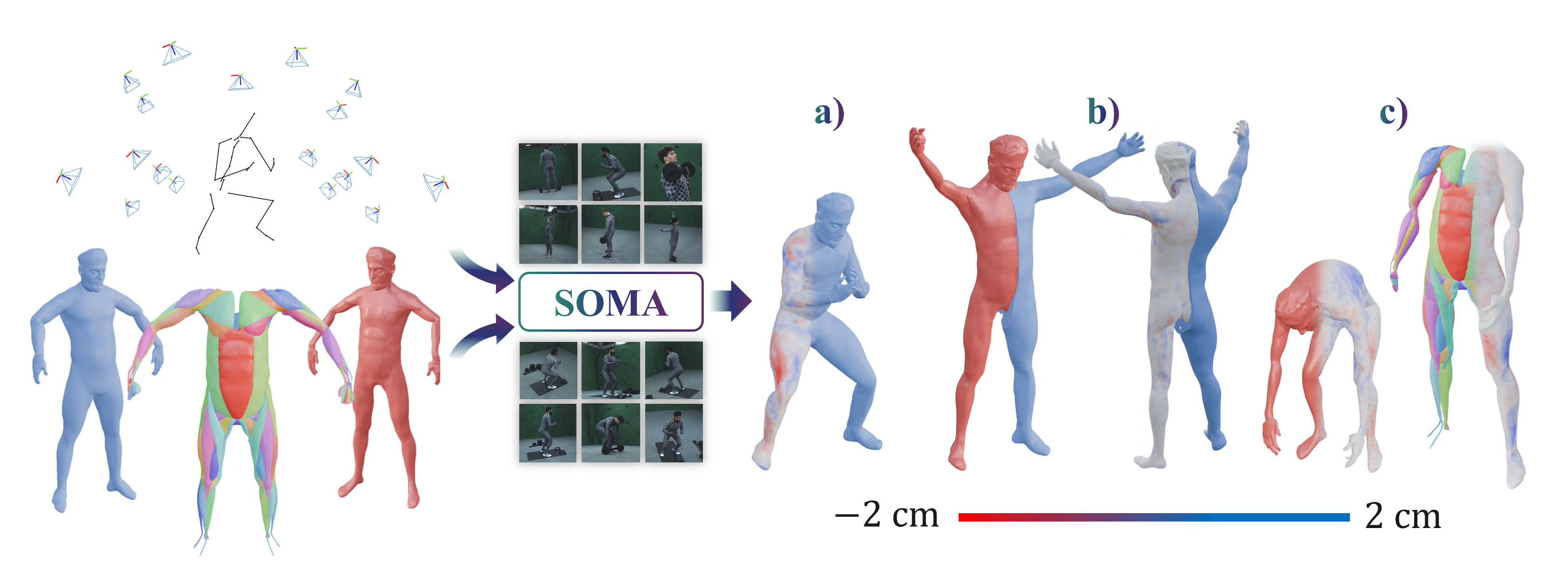}
\vspace{-5mm}
\caption{Our model \acronymtitle~allows us to convert arbitrary motion into anatomically plausible a) skin and b) muscle deformations, also targeting c) individual muscles, all grounded with real-data.}
\label{fig:teaser}
\end{figure}

\vspace{-7mm}
\begin{abstract}
    \ea{With the growing demand for realistic virtual humans, parametric body models have become a cornerstone of modern medicine, sports or entertainment applications.}
    \ea{However, most of these models are inherently limited: they only capture the 3D surface of the skin, offering no insight into the complex 
    bio-mechanical structures that generate motion.}
    \ea{As more applications expand towards biomechanics, the need for virtual human models that go beyond the skin has become increasingly evident.}
    \ea{Traditional soft-tissue simulations, such as FEM, are accurate but non-scalable and too computationally expensive for most common applications.}
    \ea{Alternatively, existing biomechanical tools can simulate muscular forces and activations, but do not model changes in external shape, restricting how activations correlate with actual observable anatomy.}
    \ea{This motivates a novel inverse research problem: recovering muscle deformations directly from visible surface observations - i.e., from the skin, and thus the pose.}
    \ea{In this work, we present \acronymtitle~(from \textbf{S}urface \textbf{O}bservations to \textbf{M}uscle \textbf{A}natomy), a person-specific model that infers spatio-temporal muscle behavior from surface signals obtained using RGB cameras, and SKIM, a subject-specific soft-tissue deformation dataset.}
    \ea{To the best of our knowledge, this is the first method that attempts to recover muscle deformations from multi-view RGB data.}
    \ea{We show how our method provides anatomically grounded animations without the complexity of traditional simulations, leading to a scalable and cost-effective solution.}
    %
    \ea{Data and code are available.}
\end{abstract}

\section{Introduction}
\label{sec:intro}


\ea{The creation of realistic and controllable digital humans is a long-standing challenge that increasingly demands biomechanical fidelity.}
\ea{Pioneering works in parametric models for human body surfaces~\cite{haslerStatisticalModelHuman2009, loperSMPLSkinnedMultiperson2015, SMPL-X:2019, osmanSTARSparseTrained2020, neumannCaptureStatisticalModeling2013, AMASS:ICCV:2019, Ramon_2023_SFLSH, softsmpl_2020, dfaust2017} have standardized the representation of body shape and pose in a low-dimensional format, enabling scalable human representations.}
%
\ea{Building on these foundations, recent vision-based approaches aim to retrieve dynamic surface geometry directly from monocular~\cite{habermannDeepCapMonocularHuman2020, habermannLiveCapRealtimeHuman2019, kocabasVIBEVideoInference2020} or multi-view RGB inputs~\cite{achenbachFastGenerationRealistic2017, liaoVINECSVideobasedNeural2023, shettyHoloportedCharactersRealtime2024, wangNeuS2FastLearning2023, zhuTriHumanRealtimeControllable2024, chen2021}, bridging the gap between parametric abstraction and real-world visual data.}
\ea{However, these models are typically based solely on surface geometry, offering no insight into the underlying internal structures.
As applications in VR, sports science, and medical simulation demand greater detail, bridging the gap between external appearance and internal physiology has become a crucial step for the next generation of digital humans.}

\ea{Current approaches to modeling internal anatomy 
have significant limitations.}
%
\ea{Heuristic methods, often used to fit anatomy to surface scans, rely on simplified assumptions that yield visually plausible but anatomically inaccurate results~\cite{kadlecekReconstructingPersonalizedAnatomical2016, gillesCreatingAnimatingSubjectSpecific2010, ichimBuildingAnimatingUserspecific2016}.}
%
\ea{Conversely, data-driven methods have 
been proven successful in solving such inverse problems and fitting anatomical structures using the human surface as input. These approaches leverage medical imaging (MRI, CT) to adapt a volumetric anatomical template, for example, for the entire body skeleton~\cite{kellerOSSOObtainingSkeletal2022, kellerSkinSkeletonBiomechanically2023, xiaReconstructingHumansBiomechanically2025} or specific regions, like the skull~\cite{achenbachMultilinearModelBidirectional2018, qiuSCULPTORSkeletonConsistentFace2022, Kadlevcek2019}. 
However, they suffer from acquisition complexity, privacy concerns, and the difficulty of capturing dynamic, full-body motion~\cite{liEfficientIncrementalPotential2023,yangLearningGeneralizedPhysical2024}, especially for methods involving soft tissue parts~\cite{komaritzanFastProjectiveSkinning2019, wenningerTailorMeSelfSupervisedLearning2024, maalinBMISelfestimatesBody2021, kellerHITEstimatingInternal2024, liNIMBLENonrigidHand2022}.}
\ea{The challenge becomes exponentially harder when we are faced with dynamic muscle behavior, due to 
its anisotropic nature and inability to obtain dynamic, full-body magnetic resonance images.}
%
\ea{Physics-based simulations can generate dynamic muscle behavior using Finite Element Methods (FEM)~\cite{modiEMUEfficientMuscle2021, muraiDynamicSkinDeformation2016, smithStableNeoHookeanFlesh2018} or Projective Dynamics~\cite{bouazizProjectiveDynamicsFusing2014, saitoComputationalBodybuildingAnatomicallybased2015}, and provide useful contributions to learning-based methods, generating synthetic data to predict muscle shape from the pose and external interactions~\cite{hanNeuralNetworkModel2024}.}

\ea{Yet, the emergence of biomechanics frameworks~\cite{sethOpenSimMusculoskeletalModeling2011} has led many methods to focus solely on predicting their mechanical behavior, also from the surface skin~\cite{chiquierMusclesAction2023, jungdamwonSIGGRAPH2025MAGNET2025, schleicherBASHBiomechanicalAnimated2021, schneiderMusclesTimeLearning2024}, thus neglecting the prediction of muscle shape.}
\ea{Consequently, there is a clear lack of scalable methods for capturing internal muscle dynamics from accessible, real-world visual data, making it difficult to build parametric models that accurately reflect internal shapes. Additionally, muscle appearance, in comparison to skeletal motion, is influenced by a broader set of factors, such as physiological cross-sectional area (PCSA) at rest, muscle activations or external interactions, all of which introduce significant complexity.}

\ea{These challenges underscore the need for a fundamentally new approach, 
one that bypasses the reliance on inaccessible medical data and instead leverages visual observations. In this work, we introduce \acronymtitle, a vision-based framework designed to recover dynamic muscle deformations directly from poses, e.g., retrieved from video (see Fig.~\ref{fig:teaser}). Our method combines surface signals with individualized anatomical templates (e.g., derived from 3D scans), enabling the recovery of rich internal representations. This allows for anatomically grounded modeling that is scalable, non-invasive, and well-suited for downstream applications.}

\ea{Our approach operates in three main stages.}
\ea{First, we introduce a vision-based data acquisition framework using a custom marker-embedded suit to capture high-resolution spatio-temporal signals of skin deformation during motion.}
\ea{Second, we propose an end-to-end data-driven pipeline that learns a non-linear blend-shape representation of a canonical muscle and skin template, supervised by these surface observations.}
\ea{Third, we integrate individual 3D muscle meshes to refine the deformation output, propagating the learned surface changes to detailed anatomical geometry.}
\ea{In summary, the contribution of this work is threefold:}
\begin{itemize}
    \item SKIM, a multi-view RGB dataset bridging surface observations and internal anatomy via high-res, spatio-temporal marker tracking and ground-truth pose.
    \item SOMA, a parametric model recovering muscle deformation from surface signals, utilizing decoupled blendshapes supervised by physics-based geometric priors.
    \item A framework propagating learned volumetric deformations to individual muscle meshes, enabling interactive internal anatomy in standard graphics engines.
\end{itemize}

\section{Related Work}
\label{sec:sota}

\ea{We review the most relevant literature to our approach across three primary domains: surface-only human body models, inverse anatomical models, and muscle-based simulations (see Tab.~\ref{tab:methods}).}

\par \textbf{Surface-only Human Body Models.}
Many parametric models 
of the body~\cite{haslerStatisticalModelHuman2009, loperSMPLSkinnedMultiperson2015, SMPL-X:2019, osmanSTARSparseTrained2020, anguelovSCAPEShapeCompletion2005, xuGHUMGHUMLGenerative2020}, hands~\cite{romeroEmbodiedHandsModeling2017}, and face~\cite{liLearningModelFacial2017} rely on standard Linear Blend Skinning (LBS) and blendshapes.
They are recovered via monocular RGB~\cite{habermannLiveCapRealtimeHuman2019, habermannDeepCapMonocularHuman2020, kanazawaEndtoEndRecoveryHuman2018, kolotourosLearningReconstruct3D2019}, multi-view setups~\cite{achenbachFastGenerationRealistic2017, liaoVINECSVideobasedNeural2023, shettyHoloportedCharactersRealtime2024, wangNeuS2FastLearning2023, zhuTriHumanRealtimeControllable2024}, or IMUs~\cite{hieuReconstructingHumanPose2024, xiaoFastHumanMotion2024}, 
achieving 
high surface fidelity. However, deformations remain purely geometric, implicitly learning soft-tissue changes without representing the underlying anatomy.

To enhance realism, some methods mimic biomechanical effects~\cite{komaritzanFastProjectiveSkinning2019} or use EMG data to predict muscle activations as 2D surface textures~\cite{chiquierMusclesAction2023, schneiderMusclesTimeLearning2024} or animation drivers~\cite{jungdamwonSIGGRAPH2025MAGNET2025}. Relying solely on visual data, Neumann et al.~\cite{neumannCaptureStatisticalModeling2013} learn data-driven surface deformations from arm muscle bulges. Ultimately, these prior works reduce musculature to abstract signals, image-space textures, or statistical surface effects. 
Because they lack explicit 3D anatomical structure, they cannot leverage true volumetric deformation properties. Our work addresses this gap by capturing and modeling explicit, subject-specific 3D muscle deformations directly from multi-view RGB body poses.

\par \textbf{Inverse Anatomical Models.}
Unlike surface-only representations, inverse models~\cite{kavanAnatomyTransfer2013, kadlecekReconstructingPersonalizedAnatomical2016, komaritzanHumansCreatingSimple2021, kellerHITEstimatingInternal2024} infer the underlying internal anatomy (e.g., bones, muscles) from external observations. This introduces the inverse challenge: recovering internal structures from sparse, often noisy surface data rather than simulating them from known biomechanical parameters.


\emph{Heuristic and Template-Based Approaches.}  
These methods non-rigidly register a generic anatomical template to a subject's surface scan. This strategy enables transferring anatomy to new body shapes~\cite{kavanAnatomyTransfer2013} and creating animation-ready models~\cite{gillesCreatingAnimatingSubjectSpecific2010, ichimBuildingAnimatingUserspecific2016}. 
While visually plausible, their reliance on heuristics and generic templates fundamentally limits their ability to faithfully reconstruct true variations in the muscles' anatomical shape, their origins, and insertions.
\begin{wraptable}{r}{0.5\linewidth}
\vspace{-8mm} 
\centering
\caption{Human Models using anatomical features. ``\mmark'' indicates methods that use real-world data but are limited to static poses or heuristics, often struggling to capture accurate dynamic motion.}
\resizebox{\linewidth}{!}{%
\begin{tabular}{l c c c}
\toprule
Method & \begin{tabular}[c]{@{}c@{}}3D Muscle \\ Representation\end{tabular} & \begin{tabular}[c]{@{}c@{}}Real-world \\ Data \end{tabular} & Full-body \\ 
\midrule
SMPL~\cite{loperSMPLSkinnedMultiperson2015}                 & \xmark & \cmark & \cmark  \\ 
Neumann et al.~\cite{neumannCaptureStatisticalModeling2013} & \xmark & \cmark & \xmark  \\
\midrule
Kavan et al.~\cite{kavanAnatomyTransfer2013}                & \cmark & \xmark & \cmark  \\
Komaritzan et al.~\cite{komaritzanHumansCreatingSimple2021} & \cmark & \mmark & \cmark  \\
Keller et al.~\cite{kellerHITEstimatingInternal2024}        & \cmark & \mmark & \cmark  \\
\midrule
Han et al.~\cite{hanNeuralNetworkModel2024}                 & \cmark & \xmark & \xmark  \\                       
Kemper et al.~\cite{botsch2026}                             & \cmark & \xmark & \cmark  \\
\midrule
\textbf{Ours}                                               & \cmark & \cmark & \cmark  \\                       
\bottomrule
\end{tabular}
}
\label{tab:methods}
\vspace{-3mm} 
\end{wraptable}


\emph{Data-driven Anatomical Reconstruction.}  
Recent data-driven methods predict subject-specific internal anatomy directly from surface shapes using paired 3D scan and medical datasets. Much focus has been on recovering skeletons~\cite{kellerOSSOObtainingSkeletal2022, xiaReconstructingHumansBiomechanically2025, kellerSkinSkeletonBiomechanically2023}, with extensions to the face/skull~\cite{achenbachMultilinearModelBidirectional2018, qiuSCULPTORSkeletonConsistentFace2022, nicolasSparseSoftDECAEfficientHighresolution2024a} and hands~\cite{liNIMBLENonrigidHand2022}. For full-body volumes, layered anatomical templates~\cite{komaritzanHumansCreatingSimple2021, wenningerTailorMeSelfSupervisedLearning2024} produce static predictions from real-world scans~\cite{maalinBMISelfestimatesBody2021} but cannot recreate dynamic results. 
Similarly, methods like HIT~\cite{kellerHITEstimatingInternal2024} learn continuous implicit representations from MRI to infer internal tissues from the outer surface. 
Since these models are trained on single, static poses, they share a fundamental limitation: they lack real-world data on how muscles dynamically deform during motion. Our work aims to address this gap.

\par \textbf{Muscle-based Simulation.}
Finite Element Method (FEM) simulations offer high accuracy by modeling volumetric biomechanical meshes~\cite{smithStableNeoHookeanFlesh2018, modiEMUEfficientMuscle2021}, enabling realistic personalized anatomy~\cite{kadlecekReconstructingPersonalizedAnatomical2016, saitoComputationalBodybuildingAnatomicallybased2015} but at expensive computational costs. To achieve real-time performance, other methods~\cite{hanNeuralNetworkModel2024} 
approximate physics by training neural networks on offline FEM data or using simplified line-actuator models like OpenSim~\cite{sethOpenSimMusculoskeletalModeling2011, schleicherBASHBiomechanicalAnimated2021}, sacrificing volumetric detail. Besides, such 
simulations struggle with reality grounding. Because dynamic real-world data (e.g., MRI~\cite{HandMRIDataset}) is scarce and mostly used for validation~\cite{blemkerThreeDimensionalRepresentationComplex2005, hicksMyModelGood2015}, simulations rely on generic physical parameters. Consequently, learning volumetric muscle deformation directly from \textit{in-vivo} motion data remains an open problem.

\section{Skin-to-Internal Muscle Dataset (SKIM)}
\label{sec:dataset}

Inferring internal muscle deformation from RGB videos is inherently ill-posed, as it relies solely on indirect skin surface observations. Resolving this requires coupling high-precision, temporally consistent skin annotations with an individualized volumetric muscle template. 

\begin{figure*}
 \centering
 \includegraphics[width=0.9\linewidth]{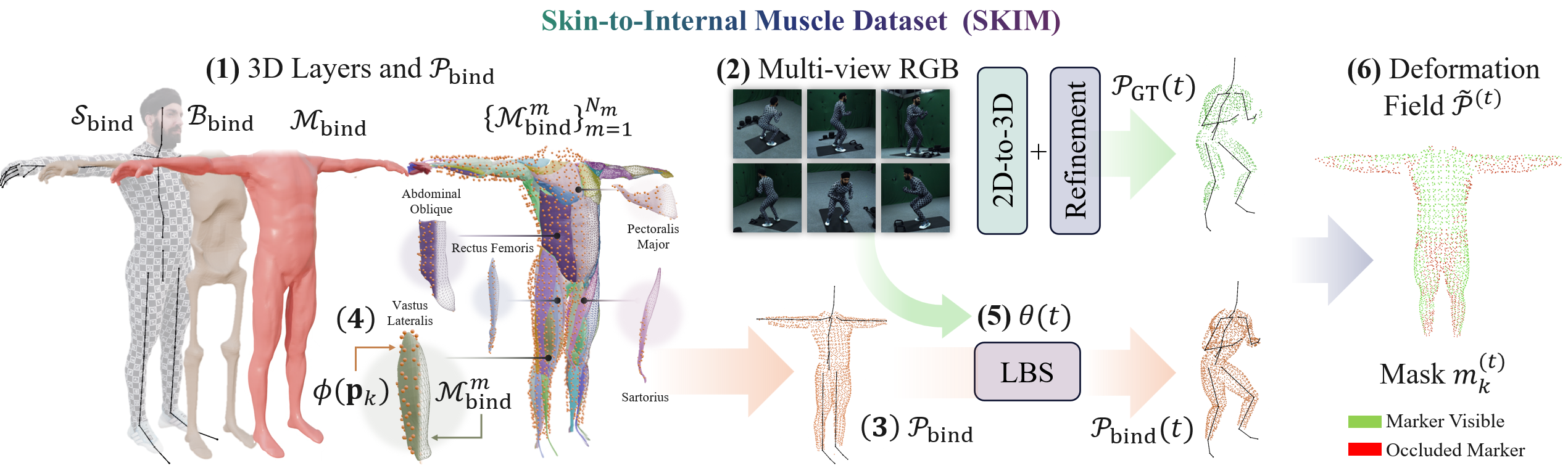}
 \caption{\textbf{SKIM Dataset Overview.} For five subjects, we provide comprehensive multi-layer data: (1) high-resolution 3D skin, bone, and individual muscle meshes; (2) 45 minutes of multi-view RGB recordings; (3) canonical ArUco marker point clouds with (4) internal muscle bindings; (5) ground-truth skeletal poses; and (6) 3D marker trajectories, residual deformation fields, and visibility masks capturing soft-tissue dynamics.}
 \label{fig:skim}
\end{figure*}

To provide these high-resolution, spatio-temporally coherent, and minimally occluded observations, we introduce the SKIM dataset (see Fig.~\ref{fig:skim}). SKIM utilizes a high-fidelity data acquisition framework centered on a custom skin-tight suit embedded with ArUco markers, captured via multi-view RGB markerless motion capture.



\vspace{-10mm}
\subsection{Capture Setup}
\label{sec:capture_setup}


We designed a custom ArUco-embedded suit~\cite{GarridoJurado2014} to capture dynamic skin deformations in our multi-view studio. This suit enables robust spatio-temporal tracking of dense surface correspondences while ensuring consistent muscle-relative placement, minimizing occlusions via known topologies, and eliminating subject-specific calibration.
\begin{wrapfigure}[14]{r}{0.3\textwidth}
  \vspace{-7mm}
  \centering
  \includegraphics[width=0.9\linewidth]{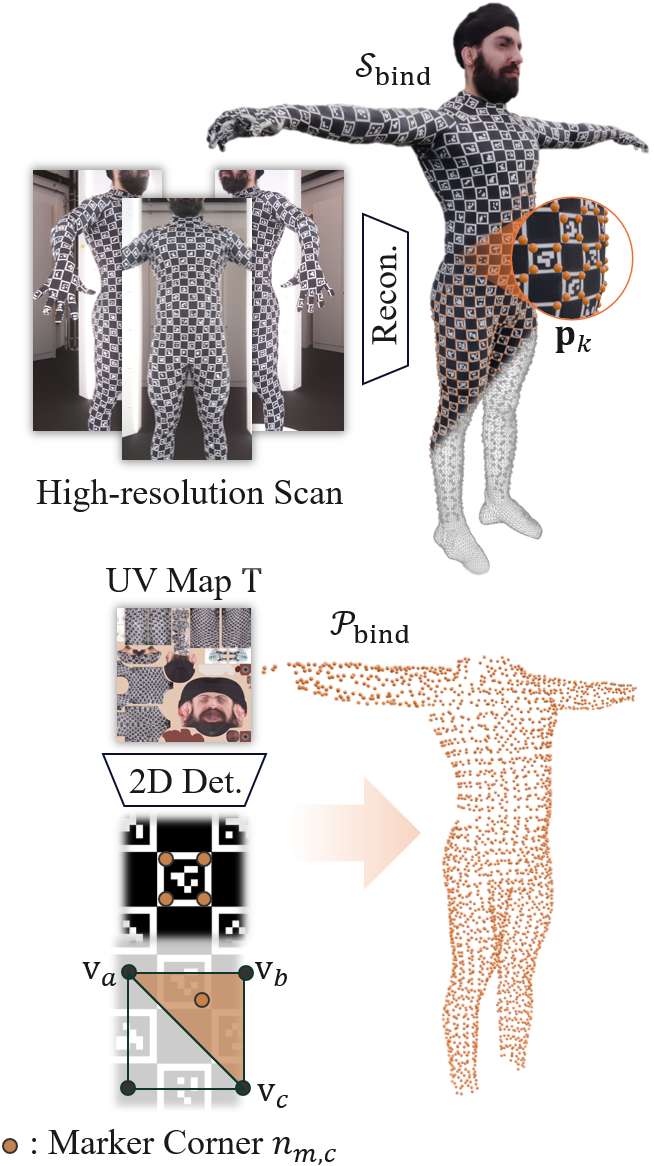}
  \caption{\textbf{Canonical Representation.}}
  \label{fig:suit}
\end{wrapfigure}
\paragraph{ArUco-based Suit Specifications.}
Constructed from stretchable, matte fabric for tight skin contact and minimal reflections, the suit's full set of landmarks is defined as $\mathcal{C} = \{ \mathbf{n}_{m,c} \mid m \in \{0, \dots, M - 1\}, c \in \{0, 1, 2, 3\} \}$, where each landmark $\mathbf{n}_{m,c}$ represents corner $c$ of the ArUco marker with dictionary ID $m$.

%
%

\paragraph{Markerless Human Capture.}
A $120$-camera mocap studio~\cite{captury} capturing at $25\,$Hz provides both the suit marker detections and the ground-truth 3D skeletal poses $\boldsymbol{\theta}_t \in \mathbb{R}^{J \times 6}$, where $J$ joints are parameterized in a continuous 6D representation~\cite{zhou2019continuity} at time $t$. A separate 140-camera scanner~\cite{Treedys} reconstructs the static body mesh templates (Sec.~\ref{sec:template}).



\subsection{Canonical Point Correspondences}
\label{sec:template}

\ea{To enable anatomically grounded modeling from 
surface observations, we first construct a canonical anatomical template that encodes the marker layout of the suit in a subject-specific 3D configuration. This template serves as a static reference point cloud $\mathcal{P}_{\text{bind}} = \{ \mathbf{p}_k \in \mathbb{R}^3 \}_{k=1}^{\mathcal{C}}$}
\ea{, where each point $\mathbf{p}_k$ is the 3D position of a specific marker corner $\mathbf{n}_{m,i,c} \in \mathcal{C}$.}%

\ea{To construct this template, we use our body scanner to generate a high-resolution static skin mesh of the subject~\cite{Achenbach2017} wearing the suit, denoted as $\mathcal{S}_{\text{bind}}$. 
The mesh $\mathcal{S}_{\text{bind}}$ consists of a set of vertices $\mathcal{V}_{\text{bind}} = \{\mathbf{v}_j \in \mathbb{R}^3 \}_{j = 1}^{V}$ and faces $\mathcal{F}_{\text{bind}} = \{\mathbf{f}_k\}_{k = 1}^{F}$, where each face $\mathbf{f}_k$ is a triangle defined by three vertex indices.}
\ea{To robustly detect marker IDs and their corner positions, we unwrap the texture of $\mathcal{S}_{\text{bind}}$ into a 2D atlas $\mathbf{T}$ 
%
%
, perform ArUco detection on $\mathbf{T}$, and re-project the detected marker corners back onto the 3D mesh.} 
\ea{During unwrapping, texture seams might occur, leading to markers being cut. The process to solve such artifacts is explained in detail in the Supplementary Material.}
\ea{For each detected corner, we identify the corresponding triangle $\mathbf{f}_k = (\mathbf{v}_a, \mathbf{v}_b, \mathbf{v}_c)$ and compute its 3D location using barycentric interpolation $\mathbf{p}_k = \lambda_1 \mathbf{v}_a + \lambda_2 \mathbf{v}_b + \lambda_3 \mathbf{v}_c, \;\text{s.t.}\; \sum \lambda_i = 1, \lambda_i \geq 0,$}
\ea{being $\mathbf{p}_k \in \mathbb{R}^3$ a point in the canonical point cloud $\mathcal{P}_{\text{bind}}$ (see Fig.~\ref{fig:suit}).}

\subsection{Internal Anatomical Modeling}
\label{sec:muscle_modeling}

\begin{figure}
 \centering
 \includegraphics[width=0.65\linewidth]{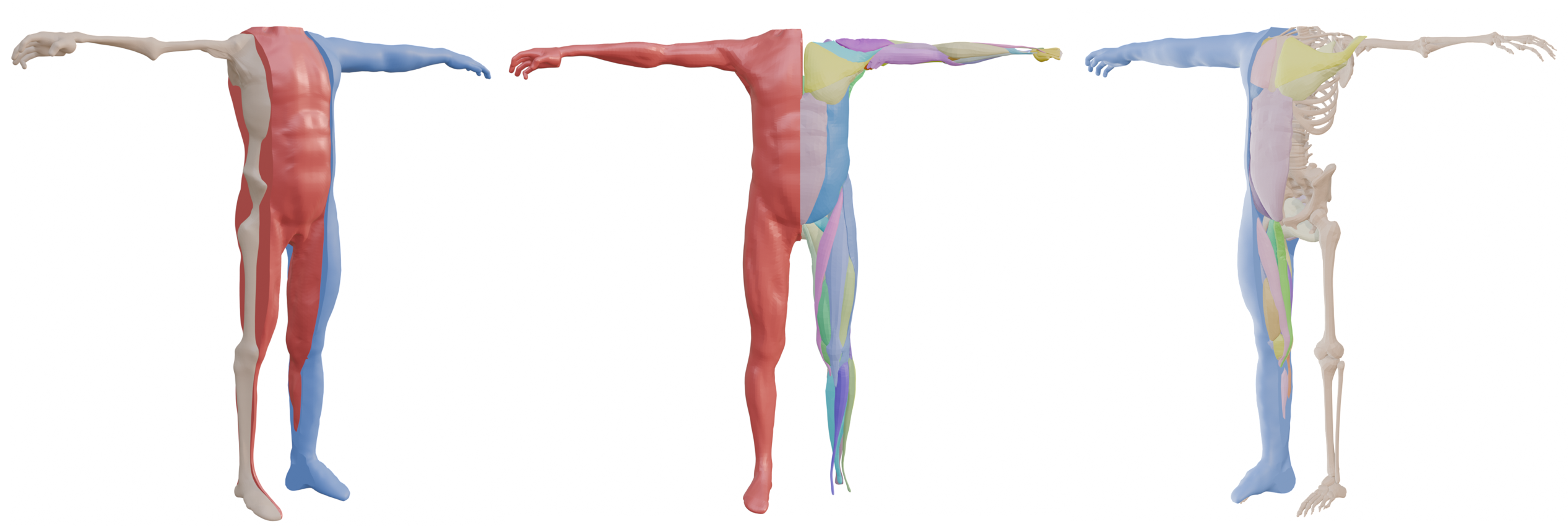}
 \caption{\textbf{Anatomical Representations.} \ea{Left: Set of bone $\mathcal{B}_{\text{bind}}$, muscle $\mathcal{M}_{\text{bind}}$ and skin layer $\mathcal{S}_{\text{bind}}$. Center: High-res muscles $\{ \mathcal{M}_{\text{bind}}^{m} \}_{m=1}^{N_{m}}$. Right: Full High-res Representations.}}
 \label{fig:layers}
\end{figure}

\ea{The internal canonical anatomy is reconstructed from the outer mesh $\mathcal{S}_{\text{bind}}$ using a volumetric fitting~\cite{komaritzanHumansCreatingSimple2021}, yielding three additional representations (see Fig.~\ref{fig:layers}):}
\begin{enumerate}
    \item A global muscle mesh $\mathcal{M}_{\text{bind}}$ that approximates the aggregated muscle volume beneath the skin. This mesh shares the same topology as $\mathcal{S}_{\text{bind}}$.
    \item A set of individualized muscle meshes $\{ \mathcal{M}_{\text{bind}}^{m} \}_{m=1}^{N_{m}}$, where each mesh represents the geometry of a specific muscle $m$ in its resting shape.
    \item A skeleton mesh $\mathcal{B}_{\text{bind}}$ that wraps the skeleton layer beneath the muscle layer. This mesh also shares the same topology as $\mathcal{S}_{\text{bind}}$.
\end{enumerate}

\vspace{-5mm}

\vspace{-5mm}
\ea{Finally, to associate each marker with its corresponding muscle, we cast a ray from the marker position $\mathbf{p}_k$ along the negative surface normal $-\alpha \mathbf{n}_k$ of its underlying triangle in $\mathcal{S}_{\text{bind}}$.
The first intersection point of this ray with the individual muscle $\mathcal{M}_{\text{bind}}^{m}$ sets the anatomical correspondence.}
%
%
This defines a mapping function $\phi(\mathbf{p}_k) = m, \;\text{s.t.}\; \mathbf{p}_k - \alpha \mathbf{n}_k \in \mathcal{M}_{\text{bind}}^{m} \;\text{and}\; \alpha > 0$,
ensuring precise marker-to-muscle correspondence. Our canonical representation of markers and muscle anatomy is shown in Fig.~\ref{fig:skim}.

\vspace{-1mm}
\subsection{Marker Tracking}
\label{sec:tracking}

Having established the (1) high-resolution skin and muscle meshes, (2) a dense set of surface markers, and (3) their specific anatomical correspondences, we now turn to leveraging our multi‑view capture system to track these markers over time.
For each frame $t$, we detect visible marker corners in the 2D image plane of each camera stream $j$, yielding detections $\mathcal{D}_t^{(j)} \subset \mathbb{R}^2$. Corners identified across multiple views are triangulated and refined by minimizing the multi-view reprojection error to find the optimal 3D coordinate $\mathbf{p}_{k,t}^*$:
\begin{equation}
\mathbf{p}_{k,t}^* = \arg\min_{\mathbf{p} \in \mathbb{R}^3} \sum_{j \in \mathcal{V}_{k,t}} \left\| \Pi_j(\mathbf{p}) - \mathbf{d}_{k,t}^{(j)} \right\|^2,
\end{equation}
where $\Pi_j$ is the camera projection operator, $\mathbf{d}_{k,t}^{(j)} \in \mathcal{D}_t^{(j)}$, and $\mathcal{V}_{k,t}$ is the set of cameras observing marker $k$ at time $t$. 
To address observation sparseness from self-occlusions, we apply post-processing strategies, such as temporal smoothing and consensus filtering, to yield refined positions $\tilde{\mathbf{p}}_{k,t}^*$. These steps are explained in detail in the Supplementary Material. The final temporally coherent ground-truth point cloud for the successfully tracked subset $\mathcal{K}_t \subset \{ 1, \dots, \mathcal{C}\}$ is defined as:
\begin{equation}
\mathcal{P}_{\text{GT}}(t) = \{ \tilde{\mathbf{p}}_{k,t}^* \mid k \in \mathcal{K}_t \}.
\end{equation}

\vspace{-1mm}
\subsection{Residual Estimation}
\label{sec:residuals}

To compare the static canonical point cloud $\mathcal{P}_{\text{bind}}$ against the temporally accurate but sparse observations $\mathcal{P}_{\text{GT}}(t)$, we articulate $\mathcal{P}_{\text{bind}}$ using the ground-truth skeletal pose $\boldsymbol{\theta}_t$ via Linear Blend Skinning (LBS). By assigning LBS weights $\mathbf{w}_k \in \mathbb{R}^J$ to each marker $\mathbf{p}_k \in \mathcal{P}_{\text{bind}}$ via barycentric interpolation from the rigged skin mesh $\mathcal{S}_{\text{bind}}$, the driven marker position is:
\begin{equation}
    \mathbf{p}_k(\boldsymbol{\theta}_t) = \sum_{j=1}^{J} w_{k,j} \left( \mathbf{R}_j(\boldsymbol{\theta}_t)\,\mathbf{p}_k + \mathbf{t}_j(\boldsymbol{\theta}_t) \right), \label{eq:lbs_marker}
\end{equation}
where $\mathbf{R}_j(\boldsymbol{\theta}_t)$ and $\mathbf{t}_j(\boldsymbol{\theta}_t)$ are the global rotation and translation of joint $j$, derived from the continuous 6D pose representation~\cite{zhou2019continuity}.
Soft-tissue dynamics (e.g., muscle bulging) cause the observed position $\mathbf{p}_{k,t}^*$ to deviate from the kinematic prediction $\mathbf{p}_k(\boldsymbol{\theta}_t)$. We isolate these non-rigid deformations as pose-corrective displacements in the canonical space by computing the world-space residual and applying the inverse blended transformation:
\begin{equation}
    \boldsymbol{\delta}_{k,t} = \left( \sum_{j=1}^{J} w_{k,j} \mathbf{R}_j(\boldsymbol{\theta}_t) \right)^{-1} \left( \mathbf{p}_{k,t}^* - \mathbf{p}_k(\boldsymbol{\theta}_t) \right). \label{eq:inverse_residual}
\end{equation}

These vectors form a sparse, temporally evolving field of canonical displacements $\mathcal{D}^{(t)} = \{ (\mathbf{p}_k, \boldsymbol{\delta}_{k,t}) \mid k \in \mathcal{K}_t \}$, serving as the ground truth for our muscle deformation model.

\vspace{-2mm}
\section{SOMA}
\label{sec:method}

\begin{figure*}
 \centering
 \includegraphics[width=1\linewidth]{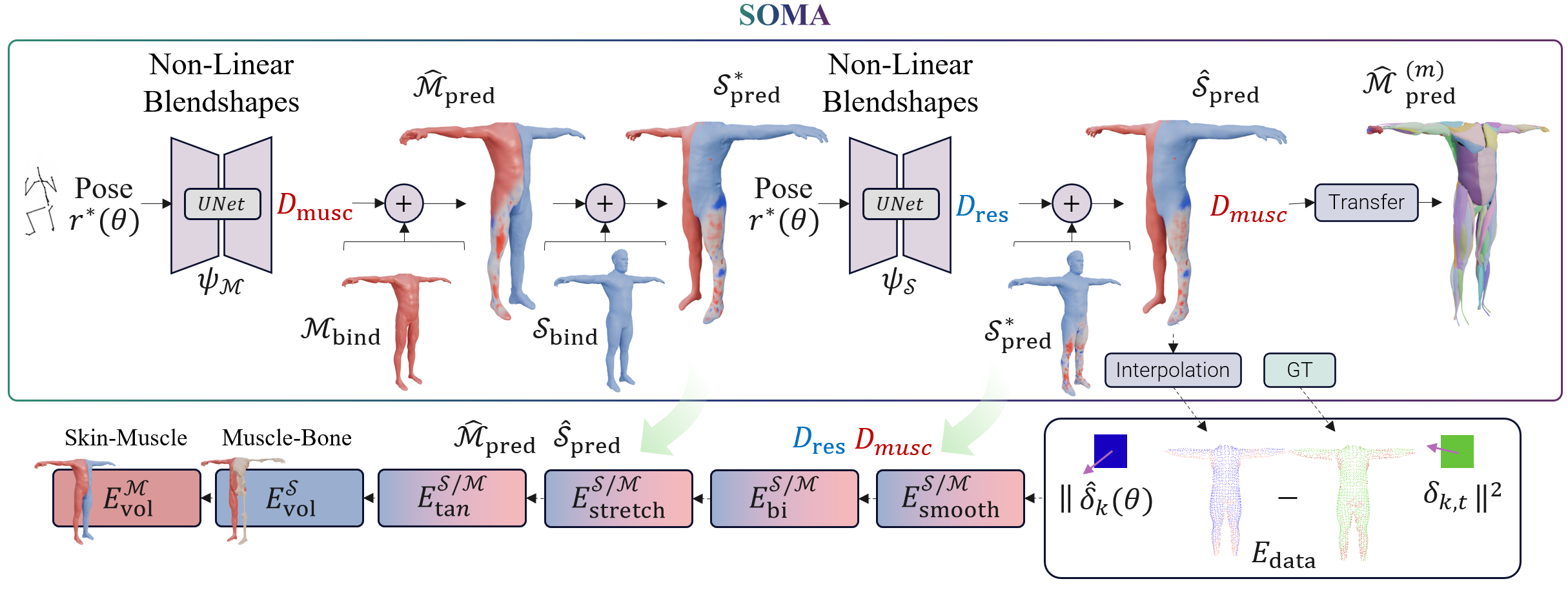}
  \caption{\textbf{Overview of SOMA.} From pose $\theta$, cascaded blendshapes predict muscle deformation ($\mathbf{D}_{musc}$) and skin residuals ($\mathbf{D}_{res}$) to reconstruct final meshes $\hat{\mathcal{M}}_{pred}$ and $\hat{\mathcal{S}}_{pred}$ from the bind geometry. To regularize marker supervision ($E_{data}$), we enforce biomechanical priors: (1) Vector constraints for smoothness ($E_{smooth}$) and sliding ($E_{tan}$); (2) Geometric constraints ($E_{stretch}$, $E_{bi}$) and soft-tissue incompressibility ($E_{vol}$).}
 \label{fig:method}
\end{figure*}

\vspace{-2mm}
\ea{Our approach builds on the concept of corrective pose-dependent blendshapes~\cite{lewis2000} but extends them to a volumetric, multi-layer anatomy representation. As we introduced in Sec.~\ref{sec:template}, let $\mathcal{M}_{\text{bind}} \in \mathbb{R}^{3N}$ denote the global muscle mesh, $\mathcal{B}_{\text{bind}} \in \mathbb{R}^{3N}$ the global skeleton mesh, and $\mathcal{S}_{\text{bind}} \in \mathbb{R}^{3N}$ the skin mesh in the rest T-pose $\boldsymbol{\theta}^*$. These layers share the same topology and skeletal rigging but represent distinct anatomical boundaries.}
\ea{Our pipeline is illustrated in Fig.~\ref{fig:method}.}

\vspace{-2mm}

\vspace{-5mm}
\subsection{Decoupled Non-Linear Blendshapes Model}

\ea{To capture complex soft tissue dynamics, we decouple deformation into interdependent structural muscle and volumetric skin layers, rather than using standard single-layer blendshapes.}

\subsubsection{Muscle Layer Deformation}

The muscle layer captures primary structural changes (e.g., muscle bulging) driven by skeletal pose $\boldsymbol{\theta}$. We employ a non-linear U-Net architecture, $\Psi_{\mathcal{M}}$, to compute a displacement field $\mathbf{D}_{\text{musc}}$ from pose features $\mathbf{r}^*(\boldsymbol{\theta}) \in \mathbb{R}^{9(J-1)}$ (defined as relative rotation matrices minus identity~\cite{loperSMPLSkinnedMultiperson2015}). The deformed canonical muscle configuration is:
\begin{equation}
\hat{\mathcal{M}}_{\text{pred}}(\boldsymbol{\theta}) 
= \mathcal{M}_{\text{bind}} + \mathbf{D}_{\text{musc}}(\boldsymbol{\theta}), \quad \text{where } \mathbf{D}_{\text{musc}} = \Psi_{\mathcal{M}}(\mathbf{r}^*).
\end{equation}

\subsubsection{Skin Layer and Residual Offset}

To allow for soft tissue compression (isochoric deformation) during articulation, the skin surface must not rigidly follow the muscle. We formulate skin deformation as a \textit{residual} function of the muscle layer:
\begin{equation}
\mathbf{D}_{\text{skin}}(\boldsymbol{\theta}) = \mathbf{D}_{\text{musc}}(\boldsymbol{\theta}) + \mathbf{D}_{\text{res}}(\boldsymbol{\theta}),
\end{equation}
where $\mathbf{D}_{\text{res}} = \Psi_{\mathcal{S}}(\mathbf{r}^*)$ is a residual offset predicted by a second U-Net. This residual provides the necessary degrees of freedom for sliding or inward compression to satisfy physical volume constraints. As this decomposition is inherently ambiguous, resolving the disentanglement is a challenge of our proposed regularization (Sec.~\ref{sec:reg}).
Finally, the canonical skin mesh $\hat{\mathcal{S}}_{\text{pred}} = \mathcal{S}_{\text{bind}} + \mathbf{D}_{\text{skin}}(\boldsymbol{\theta})$ is transformed into the final posed mesh $\mathcal{S}_{\text{pred}}$ via LBS using pose $\boldsymbol{\theta}$ and weights $\mathbf{W}$, consistent with Eq.~\ref{eq:lbs_marker}.

\subsection{Muscle and Skin Surface Supervision}
\label{sec:supervision}

We train the model end-to-end by minimizing a combined energy of data supervision and biomechanical priors: $E_{\text{total}} = \lambda_{\text{data}} E_{\text{data}} + E_{\text{reg}}$.

The data term $E_{\text{data}}$ supervises the learned corrective field using the ground-truth canonical marker residuals $\boldsymbol{\delta}_{k,t}$ (Sec.~\ref{sec:residuals}). To predict the corresponding residual $\hat{\boldsymbol{\delta}}_k(\boldsymbol{\theta})$, we interpolate the canonical skin displacement $\mathbf{D}_{\text{skin}}$ using the marker's barycentric coordinates $\mathbf{b}_k$ on its corresponding triangle vertices $\mathbf{v}_{f_j}$:
\begin{equation}
\hat{\boldsymbol{\delta}}_k(\boldsymbol{\theta}) = \sum_{j=1}^{3} b_{k,j} \mathbf{D}_{\text{skin}}(\mathbf{v}_{f_j}, \boldsymbol{\theta}).
\end{equation}

The data loss is the Mean Squared Error (MSE) between the predicted and ground-truth residuals over the $C$ total markers, masked by their visibility $m_{k,t} \in \{0, 1\}$ at frame $t$:
\begin{equation}
E_{\text{data}} = \frac{1}{N_t} \sum_{k=1}^{C} m_{k,t} \| \hat{\boldsymbol{\delta}}_k(\boldsymbol{\theta}) - \boldsymbol{\delta}_{k,t} \|^2,
\end{equation}
where $N_t$ is the number of visible markers. By computing this loss on canonical residuals rather than absolute world positions, the optimization becomes strictly invariant to global pose, forcing the network to focus solely on local non-rigid soft-tissue deformation.
By comparing residuals in the canonical frame rather than absolute world positions, the loss becomes invariant to the global pose, allowing the optimization to focus solely on the local non-rigid deformation.

\subsection{Bio-mechanically Inspired Deformation Priors}
\label{sec:reg}


Minimizing $E_{\text{data}}$ alone is ill-posed due to the disentanglement ambiguity (Sec.~\ref{sec:method}). To resolve this, we introduce biomechanical priors enforcing that muscles bulge outwards while maintaining structural stability, and skin behaves as an elastic sheet sliding over the tissue. We formulate these constraints as a combined regularization term $E_{\text{reg}}$ applied to both the muscle ($\mathcal{M}$) and skin ($\mathcal{S}$) layers:
\begin{equation}
E_{\text{reg}} = E_{\text{smooth}} + E_{\text{bi}} + E_{\text{stretch}} + E_{\text{tang}} + E_{\text{vol}}.
\end{equation}

\paragraph{Spatial Smoothness.}
To ensure smooth deformations, we apply an area-normalized Laplacian regularization. For simplicity, we refer to $\mathbf{D}_{\text{musc}}$ as $\mathbf{D}_{\mathcal{M}}$ and  $\mathbf{D}_{\text{res}}$ as $\mathbf{D}_{\mathcal{S}}$:
\begin{equation}
E_{\text{smooth}} = \sum_{l \in \{\mathcal{M}, \mathcal{S}\}} \lambda_{\text{smooth}}^{l} \| \sqrt{\mathbf{M}^{-1}} \mathbf{L}_{l} \mathbf{D}_{l}\|^2,
\end{equation}
where $\mathbf{L}_l$ is the uniform discrete Laplacian operator and $\mathbf{M}$ is the diagonal mass matrix of Voronoi areas, ensuring discretization invariance. We set $\lambda_{\text{smooth}}^{\mathcal{S}} \gg \lambda_{\text{smooth}}^{\mathcal{M}}$ to prevent high-frequency artifacts in the skin profile while allowing the underlying muscle layer to bulge freely.

\paragraph{Bending Resistance.}
To prevent localized high-frequency artifacts (e.g., isolated vertex spikes) and enforce the wide, organic deformations characteristic of thick soft tissue~\cite{Jacobson2014}, we apply a second-order Laplacian (biharmonic) regularization. Acting as a bending resistance, this is formulated as:
\begin{equation}
E_{\text{bi}} = \sum_{l \in \{\mathcal{M}, \mathcal{S}\}} \lambda_{\text{bi}}^{l} \|\mathbf{L}_{l} \mathbf{M}^{-1} (\mathbf{L}_{l} \mathbf{D}_{l})\|^2_{\mathbf{M}^{-1}}.
\end{equation}
Applying the discrete Laplacian $\mathbf{L}_l$ twice penalizes high-curvature deformations, while the inverse mass matrix $\mathbf{M}^{-1}$ maintains discretization invariance.




\paragraph{Surface Stretching.}
To preserve structural integrity and prevent unrealistic expansion, we penalize edge length deviations from their canonical rest state $L_{uv,0}$:
\begin{equation}
E_{\text{stretch}} = \sum_{l \in \{\mathcal{M}, \mathcal{S}\}} \lambda_{\text{stretch}}^{l} \sum_{(u,v) \in \mathcal{E}_l} \omega_{uv} (\| \mathbf{v}_u^l - \mathbf{v}_v^l \| - L_{uv,0})^2,
\end{equation}
where $\mathcal{E}_l$ denotes the edges of layer $l$. The weight $\omega_{uv} = (\text{area}(\mathbf{f}_i) + \text{area}(\mathbf{f}_j))/3$ assigns the area of the incident triangles $\mathbf{f}_{i,j}$ to each edge, normalized such that $\sum \omega_{uv} = 1$ to ensure scale invariance.

\paragraph{Tangential Sliding.}
To discourage unrealistic sliding, we penalize the tangential component of the displacement relative to the canonical vertex normal $\mathbf{n}_v$:
\begin{equation}
E_{\text{tang}} = \sum_{l \in \{\mathcal{M}, \mathcal{S}\}} \lambda_{\text{tang}}^{l} \sum_{v \in \mathcal{V}_l} A_v \| \mathbf{D}_{l,v} - (\mathbf{D}_{l,v} \cdot \mathbf{n}_v) \mathbf{n}_v \|^2,
\end{equation}
where $A_v$ is the local vertex area. By enforcing $\lambda_{\text{tang}}^{\mathcal{M}} \gg \lambda_{\text{tang}}^{\mathcal{S}}$, we ensure muscles remain structurally anchored (deforming primarily outwards), while allowing the skin to slide more freely over the underlying tissue during articulation.




\paragraph{Soft Tissue Volume Preservation.}

To mimic the incompressibility of human soft tissue~\cite{musclevolume2017}, we enforce volume preservation on the subcutaneous layer (skin-to-muscle, $\mathcal{S}$) and deep tissue (muscle-to-bone, $\mathcal{M}$). Modeling the space between these boundaries as sets of volumetric prisms $\mathcal{P}_l$, we penalize deviations from their rest volume $p_0$:
\begin{equation}
E_{\text{vol}} = \sum_{l \in \{\mathcal{M}, \mathcal{S}\}} \lambda_{\text{vol}}^{l} \frac{1}{|\mathcal{P}_l^*|} \sum_{p \in \mathcal{P}_l^*} \frac{\big(\text{Vol}(p) - \text{Vol}(p_0)\big)^2}{|\text{Vol}(p_0)| + \epsilon},
\end{equation}
where we evaluate only a valid subset of non-degenerate prisms $\mathcal{P}_l^*$ for numerical stability, and $\epsilon$ prevents singularities in extremely thin regions.
%
These constraints operate in different spaces: the volume ($\lambda_{\text{vol}}^{\mathcal{S}}$) is evaluated in the \textit{canonical space} and the muscle volume ($\lambda_{\text{vol}}^{\mathcal{M}}$) in the \textit{posed space},  
%
forcing the network to learn canonical muscle bulges that physically counteract LBS compression.
We compute exact signed volumes via 2-point Gauss-Legendre Quadrature. By evaluating signed rather than absolute volume, geometric inversions (layer intersections) naturally yield negative volumes and massive energy penalties, inherently preventing structural collapse. The mathematical derivation can be found in the Supplementary Material.

Finally, to animate the high-resolution individual muscles, we propagate the learned canonical displacements from the unified boundary layer $\mathcal{M}_{\text{bind}}$ to the underlying muscle meshes via a pre-computed barycentric binding mechanism, ensuring physical synchrony. The full binding and deformation mechanism is detailed in the Supplementary Material.

\section{Experiments and Results}
\label{sec:experiments}


We evaluate our muscle-driven deformation model against kinematic baselines, volumetric estimators, and ablated variations. Our analysis assesses three aspects: (1) \textbf{Data Fidelity} (tracking accuracy of ground-truth sparse markers); (2) \textbf{Anatomical Plausibility} (soft tissue volume preservation and layer interpenetration); and (3) \textbf{Geometric Quality} (surface smoothness and structural integrity).



\subsection{Data and Experimental Setup}

We evaluate our model using time-synchronized skeletal poses and tracked markers from our dataset (Sec.~\ref{sec:dataset}), testing generalization on unseen poses. Because acquiring \textit{in-vivo} volumetric ground truth during dynamic motion is infeasible, we validate physical plausibility using marker tracking and physics-inspired metrics. Specifically, we evaluate Global Surface Tracking (MPME, MedPME, P90), Dynamic Region Error (DRE) for isolating complex soft-tissue dynamics, Intersection Ratio for collision avoidance, and Volume Preservation for isochoric plausibility. We compare our explicit physical formulation against a geometric kinematic lower-bound (LBS applied to the static scan $\mathcal{S}_{\text{bind}}$) and a state-of-the-art uncoupled volumetric predictor (HIT~\cite{kellerHITEstimatingInternal2024} via SMPL~\cite{loperSMPLSkinnedMultiperson2015}). Detailed metric definitions and baseline configurations are provided in the Supplementary Material.

\subsection{Quantitative Evaluation}

Quantitative results on unseen validation sequences (see Tab.~\ref{tab:quantitative_results}) demonstrate our dual-layer formulation balances high-fidelity tracking with anatomical constraints.
While standard LBS yields a competitive global mean error (MPME) as rigid body parts heavily dilute the metric, it fundamentally fails to preserve internal structures, exhibiting volumetric joint collapse (\textit{candy-wrapper} effect). Conversely, off-the-shelf estimators like HIT~\cite{kellerHITEstimatingInternal2024} rely on generic templates lacking subject-specific proportions, resulting in high tracking errors.

Our method, \acronymtitle, resolves these issues. By capturing non-linear muscle bulging and preventing joint collapse, it improves global tracking (MedPME) and significantly reduces extreme errors (P90). Notably, in highly dynamic soft-tissue regions ($\tau > 25$ mm) where LBS degrades, our formulation cuts tracking error by more than half.
Crucially, \acronymtitle~guarantees internal physical plausibility. Standard LBS exhibits a falsely low intersection ratio (0.86\%) only because skin and deep tissue share similar skinning weights and collapse together. HIT attempts to restore volume but lacks coupled constraints, causing severe anatomical collisions (7.61\%). Only our explicitly coupled formulation successfully restores isochoric volume while strictly avoiding interpenetration.

\begin{table*}[t] 
\centering
\caption{Quantitative evaluation on unseen motion sequences. Internal anatomical plausibility is evaluated for the subcutaneous fat volume ($\mathcal{S} \rightarrow \mathcal{M}$) and muscle volume ($\mathcal{M} \rightarrow \mathcal{B}$). LBS reports no volume skin change as it does not apply pose-correctives.}
\resizebox{\textwidth}{!}{
\begin{tabular}{l c c c c c c c c c c}
\toprule
& \multicolumn{3}{c}{\textbf{Global Tracking (mm)} $\downarrow$} & \multicolumn{3}{c}{\textbf{DRE (mm)} $\downarrow$} & \multicolumn{2}{c}{\textbf{Int. (\%)} $\downarrow$} & \multicolumn{2}{c}{\textbf{$\Delta$ Vol (\%)} $\downarrow$} \\
\cmidrule(lr){2-4} \cmidrule(lr){5-7} \cmidrule(lr){8-9} \cmidrule(lr){10-11}
Method & MPME & MedPME & P90 & $>15$ & $>20$ & $>25$ & $\mathcal{S} \rightarrow \mathcal{M}$ & $\mathcal{M} \rightarrow \mathcal{B}$ & $\mathcal{S} \rightarrow \mathcal{M}$ & $\mathcal{M} \rightarrow \mathcal{B}$ \\
\midrule
LBS & \textbf{13.11} & 15.64 & 26.03 & 40.46 & 88.25 & 134.28 & 0.85 & 0.93 & N/A & 18.81 \\
HIT~\cite{kellerHITEstimatingInternal2024} & 115.34 & 50.15 & 354.80 & \textbf{24.09} & 119.72 & 151.65 & 7.61 & N/A & 2.99 & \textbf{1.95} \\
\midrule
\textbf{Ours} & 13.43 & \textbf{11.51} & \textbf{19.20} & 25.51 & \textbf{46.99} & \textbf{64.52} & \textbf{0.85} & \textbf{0.93} & \textbf{1.37} & 10.35 \\
\bottomrule
\end{tabular}
}
\label{tab:quantitative_results}
\end{table*}


\subsection{Qualitative Evaluation}

\begin{figure*}[t]
 \centering
 \includegraphics[width=0.9\linewidth]{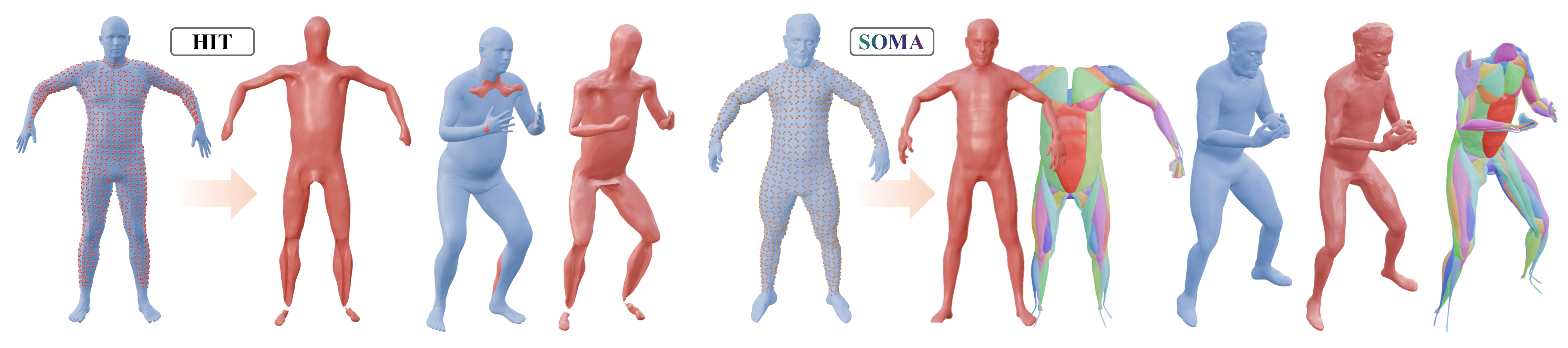}
 \caption{\textbf{Qualitative Comparison against HIT~\cite{kellerHITEstimatingInternal2024}.} Our explicit model (right) predicts finer-grained dynamic deformations. For cross-topology evaluation (scan vs. SMPL), we map our marker representation onto HIT's meshes via barycentric interpolation.}
 \label{fig:comparison_hit}
\end{figure*}

\begin{figure*}
 \centering
 \includegraphics[width=0.9\linewidth]{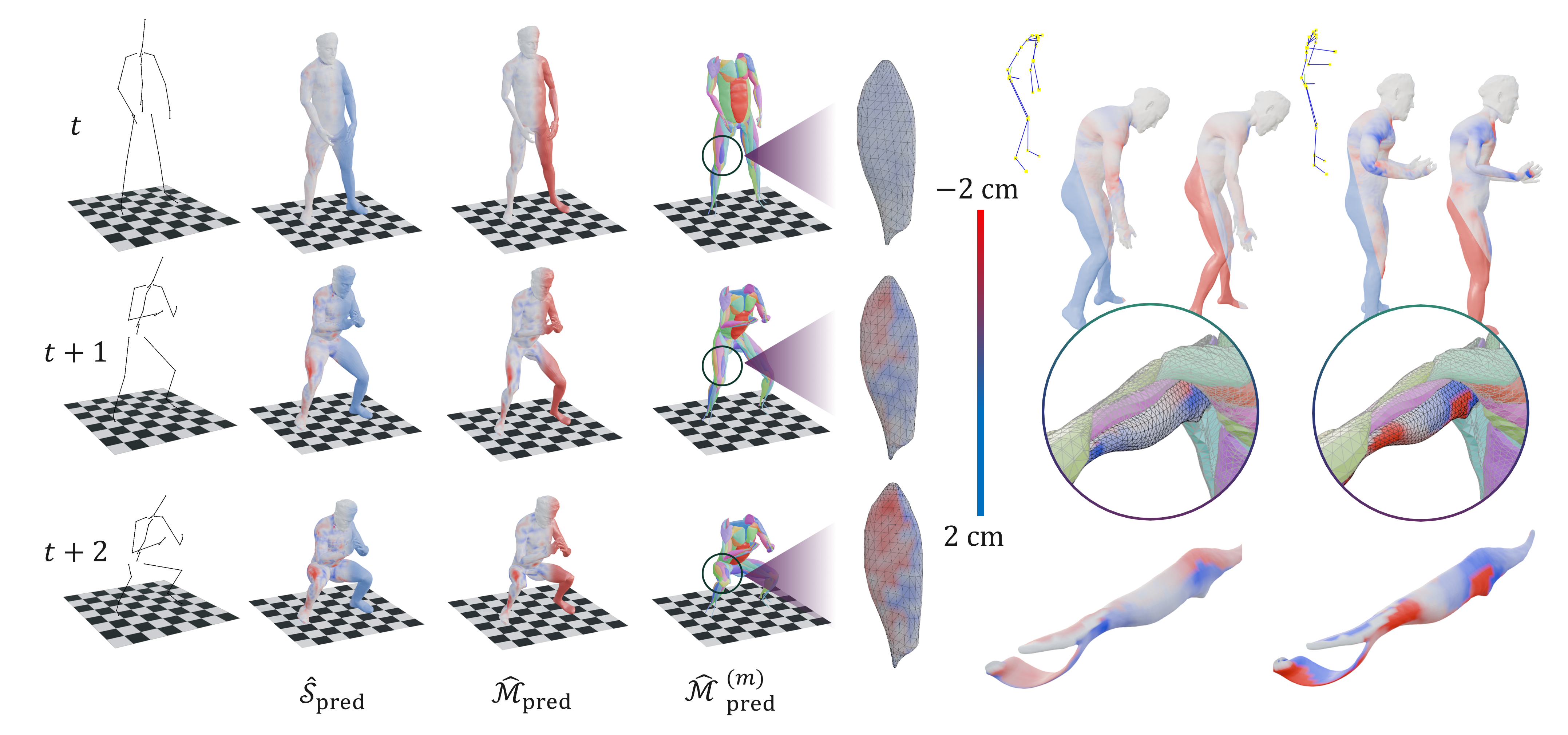}
 \caption{\textbf{From skin to individual muscle deformation.} Given a target pose $\boldsymbol{\theta}$, we visualize the inferred non-linear displacements of the skin and muscle layers (blue: inward compression, red: outward expansion).}
 \label{fig:viser}
\end{figure*}

We provide a real-time visualization tool built in Viser~\cite{yi2025viser} (as shown in Fig.~\ref{fig:viser}). By importing the trained model and subject bind meshes ($\mathcal{M}_{\text{bind}}$, $\mathcal{S}_{\text{bind}}$), the tool enables interactive full-body inference, per-muscle querying, and isolated single-muscle prediction using only locally associated markers, therefore capturing active muscle bulging (e.g., quadriceps, triceps) during motion. 
We also compare dynamic deformations against HIT~\cite{kellerHITEstimatingInternal2024} (see Fig.~\ref{fig:comparison_hit}). While static baselines like these produce plausible resting volumes, they over-smooth high-frequency dynamics. 
For more example motions, please see the Supplementary Material.



\vspace{0mm}

\vspace{0mm}


\subsection{Ablations}

We ablate our architectural choices and loss components across three categories: \textbf{network architecture}, \textbf{vector regularization} (on displacement fields), and \textbf{physical energies} (on surface geometry).

\paragraph{Network Architecture.}
We compare Linear Blendshapes, an MLP, and a U-Net for predicting canonical displacements in Tab.~\ref{tab:ablation_architecture}. Although similar, linear models struggle more with non-linear soft-tissue bulging and lack the spatial weight-sharing needed to enforce global volume constraints. Though MLPs add non-linearities, the U-Net best captures both local surface details and spatial deformations without overfitting the sparse marker data.

\paragraph{Vector Regularization ($E_{\text{smooth}}, E_{\text{tan}}$).}
Removing these terms (\textit{w/o Vector}) surprisingly degrades subcutaneous volume preservation (see Tab.~\ref{tab:ablations_energies}). Without tangential constraints, vertices drift laterally to minimize marker error, shearing the underlying prisms.  This prevents $E_{\text{vol}}$ from enforcing outward biological bulging and slightly degrades global tracking.

\paragraph{Physical Energies ($E_{\text{bi}}, E_{\text{stretch}}, E_{\text{vol}}$).}
Our \textit{Full Model} achieves the lowest tracking error (13.43 mm), demonstrating that these geometric priors act as structural anchors guiding the network to better local minima. Removing all physical constraints (\textit{w/o Physics}) yields the worst tracking and high volume distortion. Ablating structural tension (\textit{w/o $E_{\text{bi}}, E_{\text{stretch}}$}) causes the highest volume errors, as the mesh artificially satisfies volume constraints via severe high-frequency wrinkling instead of smooth bulging. Ablating $E_{\text{vol}}$ spikes subcutaneous volume error to 8.28\%, confirming its necessity in correcting LBS artifacts. Finally, while the full formulation reduces outer fat volume error to 1.37\%, the deep muscle layer ($\mathcal{M} \rightarrow \mathcal{B}$) retains a 10.35\% deviation. This represents a realistic physical compromise: the optimizer prioritizes strict non-intersection over full deep-tissue inflation to prevent massive skeletal collisions.

\begin{table*}[t]
\centering
\begin{minipage}[t]{0.48\textwidth}
\centering
\caption{Ablation of network architectures. While the U-Net achieves the lowest error, Linear and MLPs are sufficient to effectively fit the markers. This demonstrates the significant benefit of training person-specific models on high-quality tracking data, while highlighting the inherent anatomical misalignment issues faced by generic templates.}
\resizebox{\linewidth}{!}{%
\begin{tabular}{l c c c c c c}
\toprule
& \textbf{Global} $\downarrow$ & \textbf{DRE} $\downarrow$ & \multicolumn{2}{c}{\textbf{Int. (\%)}} & \multicolumn{2}{c}{\textbf{$\Delta$ Vol (\%)}} \\
\cmidrule(lr){2-2} \cmidrule(lr){3-3} \cmidrule(lr){4-5} \cmidrule(lr){6-7}
Architecture & MPME & $>25$ & $\mathcal{S} \rightarrow \mathcal{M}$ & $\mathcal{M} \rightarrow \mathcal{B}$ & $\mathcal{S} \rightarrow \mathcal{M}$ & $\mathcal{M} \rightarrow \mathcal{B}$ \\
\midrule
Linear       & 16.52 & 64.58 & 0.85 & 0.93 & 3.57 &  11.02 \\
MLP          & 14.03 & 64.93 & 0.86 & 1.09 & 5.74 & 11.05  \\
\midrule
\textbf{U-Net} & \textbf{13.43} & \textbf{64.52} & \textbf{0.85} & \textbf{0.93} & \textbf{1.37} & \textbf{10.35} \\
\bottomrule
\end{tabular}%
}
\label{tab:ablation_architecture}
\end{minipage}\hfill
\begin{minipage}[t]{0.48\textwidth}
\centering
\caption{Ablation of Regularization Terms. While all priors significantly improve anatomical stability, the smoothness and stretch energies ($E_{\text{bi, str}}$) prove most critical. Removing them causes the most severe volumetric distortion, highlighting that membrane tension is essential to prevent mesh artifacts during dynamic motion.}
\resizebox{\linewidth}{!}{%
\begin{tabular}{l c c c c c c}
\toprule
& \textbf{Global} $\downarrow$ & \textbf{DRE} $\downarrow$ & \multicolumn{2}{c}{\textbf{Int. (\%)}} & \multicolumn{2}{c}{\textbf{$\Delta$ Vol (\%)}} \\
\cmidrule(lr){2-2} \cmidrule(lr){3-3} \cmidrule(lr){4-5} \cmidrule(lr){6-7}
Method & MPME & $> 25$ & $\mathcal{S} \rightarrow \mathcal{M}$ & $\mathcal{M} \rightarrow \mathcal{B}$ & $\mathcal{S} \rightarrow \mathcal{M}$ & $\mathcal{M} \rightarrow \mathcal{B}$ \\
\midrule
w/o Physics & 13.71 & 64.93 & 0.87 & 0.95 & 9.37 & 13.54 \\
w/o Vector  & 13.46 & 64.57 & 0.87 & 0.94 & 8.27 & 12.63 \\
\midrule
w/o $E_{\text{bi, str}}$ & 13.54 & 64.48 & 0.88 & 0.94 & 10.54 & 13.76 \\
w/o $E_{\text{vol}}$     & 13.67 & 64.57 & 0.87 & 0.97 & 8.28 & 11.87 \\
\midrule
\textbf{Full (Ours)} & \textbf{13.43} & \textbf{64.52} & \textbf{0.85} & \textbf{0.93} & \textbf{1.37} & \textbf{10.35} \\
\bottomrule
\end{tabular}%
}
\label{tab:ablations_energies}
\end{minipage}
\vspace{-4mm}
\end{table*}
\section{Limitations and Future Work}
\label{sec:limitations}
While our proposed method is among the first to recover interior muscle deformations solely from visual data, it still has some limitations. 
First, we rely on volumetric fitting~\cite{komaritzanHumansCreatingSimple2021} for the initial template, which makes us vulnerable to initialization errors; future work could jointly optimize baseline anatomy and dynamic displacements. 
Second, requiring a marker suit restricts convenience. Leveraging our data to train robust, markerless skin trackers would be a promising alternative. 
Third, force- and co-contraction-driven variation at a fixed joint angle constitutes a remaining unmodeled component.
Finally, our method is currently subject-specific. Expanding the dataset to build a more generalizable parametric model offers an exciting path toward in-the-wild anatomically accurate performance capture.

%
%
\section{Conclusion}
\label{sec:conclusion}
We introduced SOMA, a novel approach that models skin and muscle geometry driven by skeletal motion. To achieve this, we developed a capture setup for high-fidelity spatio-temporal surface tracking. As learning internal anatomy from visual data is inherently ill-posed, our blendshapes formulation leverages biomechanically-inspired priors. At inference, SOMA animates both layers using only user-defined skeletal poses. We believe this work is a crucial step toward accessible, data-driven biomechanical digital humans, inspiring future work in the directions of human surface correspondence estimation and person-agnostic anatomical human models.


\section{Acknowledgements}
This research has been funded by the Ministry of Culture and Science of the State of North Rhine-Westphalia through the project inVirtuo 4.0 (PB22-063-B) and by the Federal Ministry of Education and Research of Germany and the state of North Rhine-Westphalia as part of the Lamarr Institute for Machine Learning and Artificial Intelligence.

\clearpage
%
%
\bibliographystyle{splncs04}
\bibliography{biblio}
\end{document}


\title{SOMA: From Surface Observations to Muscle Anatomy} 

\titlerunning{SOMA}


\author{Eduardo Alvarado\inst{1}\orcidlink{0000-0003-3395-5674}\and
Emily Kim\inst{1}\and
Gerrit Nolte\inst{2}\orcidlink{0000-0002-5080-1039}\and
Friedemann Runte\inst{2}\orcidlink{0009-0000-1963-8707}\and
Mario Botsch\inst{2}\orcidlink{0000-0001-9954-120X}\and
Marc Habermann\inst{1}\orcidlink{0000-0003-3899-7515}\and
Christian Theobalt\inst{1}\orcidlink{0000-0001-6104-6625}}

\authorrunning{E.~Alvarado et al.}

\institute{
Max Planck Institute for Informatics, Saarland Informatics Campus \\
\and
TU Dortmund University \\
[4pt]
\url{https://vcai.mpi-inf.mpg.de/projects/SOMA/}
}

\maketitle

\section{SKIM Dataset Details}
\label{sec:supp_dataset}

\ea{In this section, we provide a detailed breakdown of the SKIM dataset, including suit manufacturing, subject demographics, the specific motion sequences captured, and the specifications of the anatomical templates.}

\begin{figure*}
 \centering
 \includegraphics[width=1\linewidth]{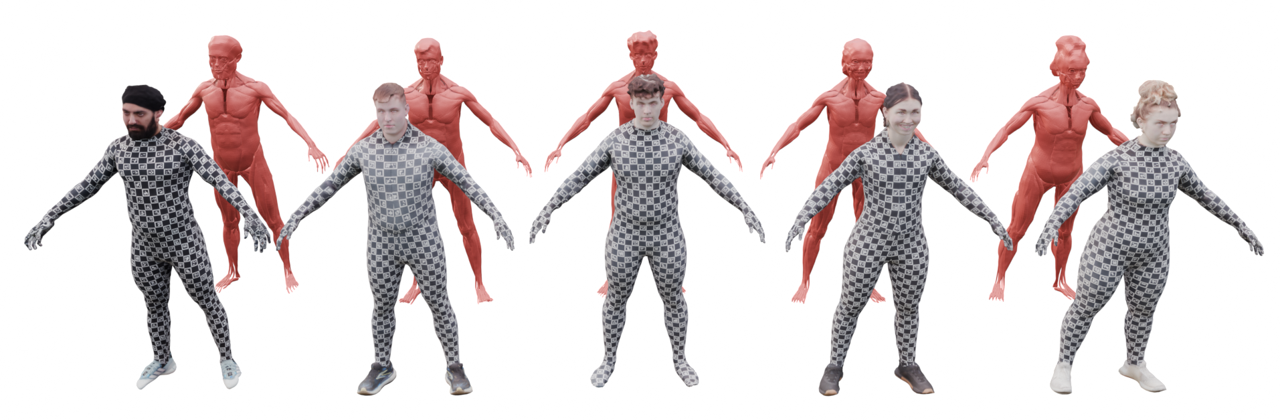}
 \caption{\textbf{SKIM.} Our recorded subjects (S1 to S5).}
 \label{fig:subjects}
\end{figure*}

\subsection{Suit Design and Marker Encoding}
\label{sec:supp_suit}

The motion capture suit is custom-manufactured using a stretchable, matte fabric to minimize specular reflections and ensure tight contact with the skin surface. To capture high-frequency deformations across the entire body, the suit requires a dense marker layout. It contains a total of $4$x$4$ $2310$ ArUco markers~\cite{GarridoJurado2014}. However, the standard ArUco dictionary is limited to a maximum of $M=1000$ unique IDs. To achieve the required density, we employ a multi-instance encoding scheme where specific IDs might be repeated once. We distinguish between duplicates using an instance index $i \in \{0, 1\}$.
Formally, the full set of landmarks $\mathcal{C}$ on the suit is defined as:
\begin{equation}
\mathcal{C} = \{ \mathbf{n}_{m,i,c} \mid m \in \{0, \dots, M - 1\}, i \in \{0, 1\}, c \in \{0, 1, 2, 3\} \},     
\end{equation}
where each landmark $\mathbf{n}$ is a specific marker corner defined by the dictionary ID $m$, the instance index $i$, and the corner index $c$. This encoding allows us to map the $2310$ physical markers to a unique 3D location in the canonical template.

\subsection{Subject Demographics and Templates}
\ea{SKIM contains data from 5 subjects (3 Male, 2 Female) (see Fig.~\ref{fig:subjects}) with varying body shapes and BMIs to ensure anatomical diversity. Tab.~\ref{tab:demographics} summarizes the physical attributes of each participant.}
\ea{As described in the main paper, we generate individualized volumetric templates for each subject. Tab.~\ref{tab:mesh_specs} details the resolution of these meshes, which are shared across all subjects.}

\begin{table*}[t]
\centering
\begin{minipage}[t]{0.48\textwidth}
\centering
\caption{\textbf{Subject demographics.} \ea{Physical attributes of the 5 subjects in SKIM.}}
\begin{tabular*}{\linewidth}{@{\extracolsep{\fill}}lcccc@{}}
\toprule
ID & Sex & \begin{tabular}{c}Height\\(cm)\end{tabular} &
\begin{tabular}{c}Weight\\(kg)\end{tabular} & BMI \\
\midrule
S1 & M & 178 & 82 & 25.9 \\
S2 & M & 182 & 92 & 27.8 \\
S3 & M & 181 & 90 & 27.5 \\
S4 & F & 168 & 81 & 28.7 \\
S5 & F & 162 & 60 & 23.1 \\
\bottomrule
\end{tabular*}
\label{tab:demographics}
\end{minipage}\hfill
\begin{minipage}[t]{0.48\textwidth}
\centering
\caption{\textbf{Mesh Specifications.} \ea{Vertex and face counts for the canonical templates.}}
\begin{tabular*}{\linewidth}{@{\extracolsep{\fill}} p{0.5\linewidth} c c @{}}
\toprule
\textbf{Component} & \textbf{Vertices} & \textbf{Faces} \\
\midrule
Skin Bind $\mathcal{S}_{\text{bind}}$ & \ea{23,752} & \ea{47,500} \\
Muscle Bind $\mathcal{M}_{\text{bind}}$ & \ea{23,752} & \ea{47,500} \\
Bone Bind $\mathcal{B}_{\text{bind}}$ & \ea{23,752} & \ea{47,500} \\
\midrule
Individual Muscles $\{\mathcal{M}_{\text{bind}}^{m}\}$ (Total) & \ea{320,256} & \ea{118,758} \\
\bottomrule
\end{tabular*}
\label{tab:mesh_specs}
\end{minipage}
\end{table*}


\subsection{Acquisition Protocol}
\ea{Each subject underwent a capture session lasting approximately 45 minutes, while they were equipped with our custom-fit suit embedded with markers.}
\ea{The session was designed to capture how different muscles change by observing the skin surface during intense physical exertion.}
\ea{The protocol was divided into three main distinct phases:}
\begin{enumerate}
    \item \textbf{Part I: Extended Warm-Up \& Calibration.}
    \ea{Subjects performed a standardized 10-minute routine including walking, jogging, and dynamic body-weight exercises (squats, lunges, push-ups) to mobilize joints and establish a physiological baseline. This phase concluded with arbitrary stretching.}
    \item \textbf{Part II: Weighted Resistance Training.}
    \ea{The core of the dataset targets key muscle groups (i.e., arms, shoulders, back, legs) using dumbbells and barbells. Subjects performed two rounds of five specific exercises (see Tab.~\ref{tab:motions}). The first round utilized lighter weights, while the second employed increased loads, encouraging muscle bulging and facilitate capture.} 
    \item \textbf{Part III: Post-exercise Comparison.}
    \ea{Immediately following the resistance training, subjects repeated the initial warm-up routine, allowing for a direct comparison of surface geometry before and after exercise, capturing effects such as muscle pumping (increased volume due to blood flow).}
\end{enumerate}

\begin{table}[h]
\centering
\caption{\textbf{Motion Sequences.} \ea{Standardized exercises performed by all subjects.}}
\label{tab:motions}
\begin{tabular}{l l l}
\toprule
\textbf{Category} & \textbf{Motion} & \textbf{Muscle Focus} \\
\midrule
\multirow{4}{*}{Warm-Up}
& Walking / Jogging & Locomotion \\
& Bodyweight Squats & Quadriceps / Glutes \\
& Lunges & Legs / Balance\\
& Push-ups & Pectorals / Triceps\\
\midrule
\multirow{5}{*}{Routine}
& Dumbbell Bicep Curls & Biceps Brachii \\
& Overhead Tricep Extension & Triceps Brachii \\
& Standing Military Press & Deltoids / Shoulders \\
& Barbell Bent-Over Rows & Upper Back / Lats \\
& Barbell Deadlifts & Posterior Chain \\
\bottomrule
\end{tabular}
\end{table}

\subsection{ArUco Annotation}

As described in the main paper (Sec. 3.2), we establish the canonical marker template $\mathcal{P}_{\text{bind}}$ by detecting ArUco markers on the unwrapped 2D texture atlas $\mathcal{T}$ of the bind shape $\mathcal{S}_{\text{bind}}$. While standard ArUco detection libraries are robust, ensuring a complete set of markers is challenging due to two primary factors: (1) fail detections due to local distortions introduced by the UV unwrapping process or varying lighting conditions, and (2) texture seams, as unwrapping the 3D body into a 2D plane requires cuts (seams) and markers can be physically located on these boundaries. Standard algorithms, which expect contiguous quadrilaterals, inherently fail to detect these markers.
\begin{wrapfigure}[10]{r}{0.5\linewidth}
    \centering
    \vspace{-9mm}
    \includegraphics[width=\linewidth]{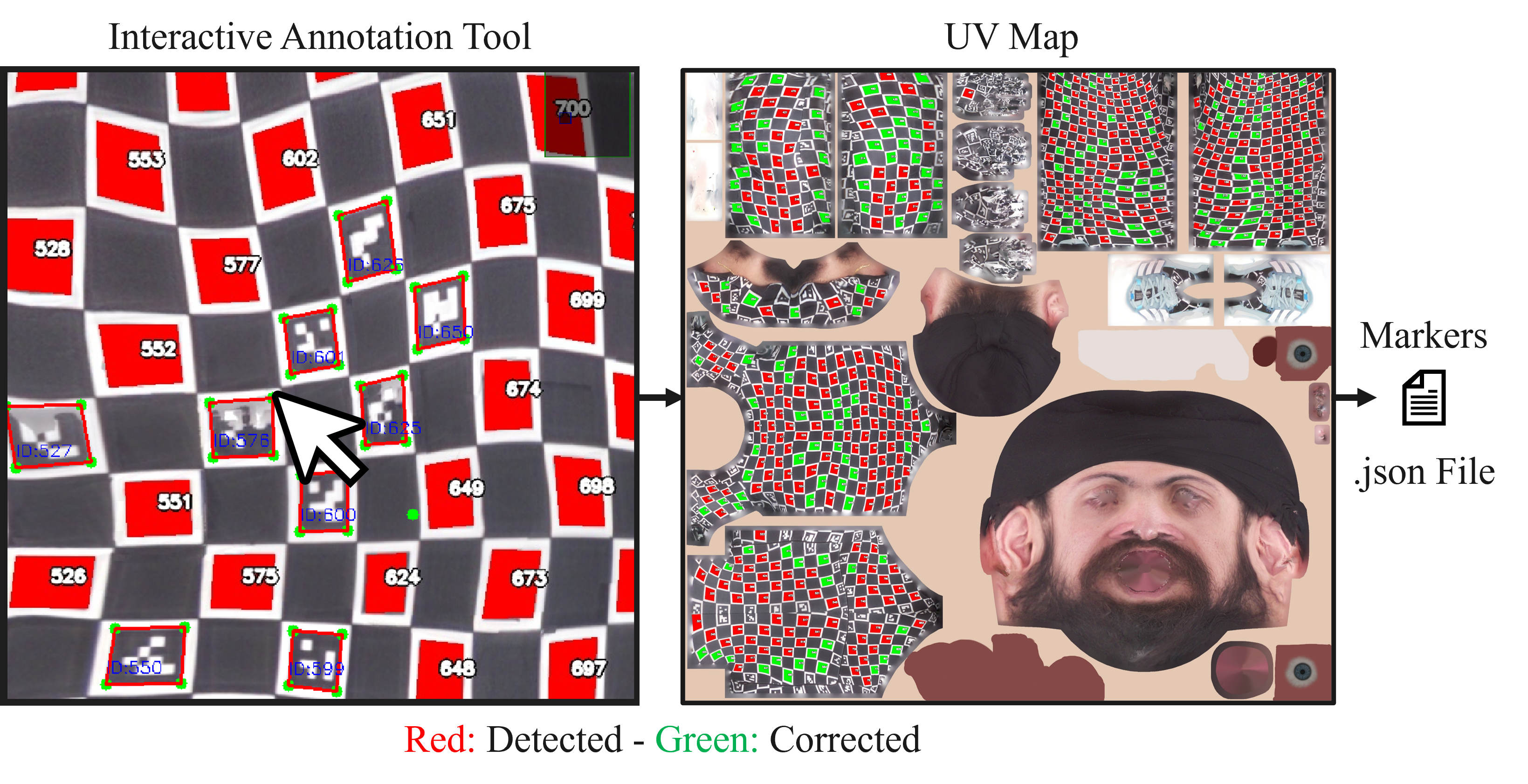}
    \caption{\textbf{Interactive Annotation Tool.}}
    \label{fig:annotation}
\end{wrapfigure}
To overcome these limitations, we developed a custom interactive annotation tool to visualize the UV map $\mathcal{T}$ and overlay the successfully detected markers (see Fig.~\ref{fig:annotation}). It enables the user to manually annotate missing markers by selecting their four corners directly on the image. The tool is designed to handle seam artifacts: it allows the user to select corners that are spatially separated in the 2D UV map (e.g., two corners on one UV island and two on another) and groups them under a single, consistent Marker ID. The system automatically validates that each ID is associated with exactly four corner coordinates and merges the manual annotations with the automatic detections. This process yields the final corrected JSON structure required to build the dense 3D canonical point cloud $\mathcal{P}_{\text{bind}}$ via barycentric interpolation.


\newpage
\subsection{Data Structure and Size}
\ea{The SKIM dataset is organized by subject (`S1` to `S5`). For each subject, we provide the canonical anatomical definition, the specific layer meshes, the raw sequence data, and a pre-processed version optimized for training:}

\begin{itemize}[nosep,leftmargin=*]
    \item \texttt{/S[1-5]/}
    \begin{itemize}[nosep,leftmargin=*]
        
        \item \texttt{/canonical\_model/} – \ea{Data defining the subject's bind configuration.}
        \begin{itemize}[nosep,leftmargin=*]
            \item \texttt{canonical\_markers\_tpose.json} – \ea{Canonical point-cloud $\mathcal{P}_{\text{bind}}$ in T-Pose.}
            \item \texttt{lbs\_weights\_markers.json} – \ea{Weights $\mathbf{w}_k$ binding markers to skeleton.}
            \item \texttt{lbs\_weights\_skin.json} – \ea{Weights binding the skin mesh $\mathcal{S}_{\text{bind}}$ to the skeleton, shared across $\mathcal{M}_{\text{bind}}$ and $\mathcal{B}_{\text{bind}}$.}
        \end{itemize}

        \item \texttt{/layers/} – \ea{High-resolution anatomical templates in bind pose.}
        \begin{itemize}[nosep,leftmargin=*]
            \item \texttt{skin\_bind.obj} – \ea{Surface scan mesh $\mathcal{S}_{\text{bind}}$.}
            \item \texttt{muscle\_bind.obj} – \ea{Global volumetric muscle mesh $\mathcal{M}_{\text{bind}}$.}
            \item \texttt{bone\_bind.obj} – \ea{Global volumetric bone mesh $\mathcal{B}_{\text{bind}}$.}
            \item \texttt{/individual\_muscles/} – \ea{Directory with individual muscle meshes $\{\mathcal{M}_{\text{bind}}^{m}\}$.}
        \end{itemize}

        \item \texttt{/raw/} – \ea{Captured motion sequences.}
        \begin{itemize}[nosep,leftmargin=*]
            \item \texttt{/shot\_XXX/}
            \begin{itemize}[nosep,leftmargin=*]
                \item \texttt{motion.bvh} – \ea{GT skeletal motion $\boldsymbol{\theta}_t$.}
                \item \texttt{residuals.json} – \ea{Surface residuals $\boldsymbol{\delta}_{k,t}$.}
                \item \texttt{masks.json} – \ea{Visibility masks $m_{k,t}$.}
            \end{itemize}
        \end{itemize}

        \item \texttt{/processed/} – \ea{Data pre-processed for training.}
        \begin{itemize}[nosep,leftmargin=*]
            \item \texttt{shot\_XXX\_frame\_XXXX.npy} – \ea{Consolidated frame-based state vectors (pose, residuals, masks).}
        \end{itemize}

    \end{itemize}
\end{itemize}

\vspace{4mm}
Each of the 5 subjects was recorded for approximately 45 minutes at 25 Hz, in a motion capture studio equipped with 120 synchronized cameras.
\begin{wraptable}{r}{0.50\textwidth}
\centering
\vspace{-7mm}
\caption{Dataset Size Breakdown.}
\label{tab:dataset_size}
\resizebox{0.40\textwidth}{!}{%
\begin{tabular}{l c c c}
\toprule
\textbf{Subject} & \textbf{\shortstack{Duration \\ (min)}} &
\textbf{\shortstack{2D Detected \\ Frames}} &
\textbf{\shortstack{3D Detected \\ Frames}} \\
\midrule
S1 & $\approx 45$ & $\approx 4{,}76$M & $\approx 50$K \\
S2--S5 (each) & $\approx 45$ & $\approx 7{,}24$M & $\approx 54$K \\
\midrule
\textbf{Total} & \textbf{225} & \textbf{33{,}7M} & \textbf{266K} \\
\bottomrule
\end{tabular}
} 
\end{wraptable}
This results in over 330,000 unique time steps of motion and nearly 40 million individual 2D images processed for marker detection. Tab.~\ref{tab:dataset_size} provides a detailed breakdown of the recorded data volume per subject.

\section{Marker Refinement}
\label{sec:refinement}

\ea{The observed point-cloud across the motion sequence $t \in \{1, \dots, T\}$ is defined as:}
\begin{equation}
    \mathcal{P}_{\text{GT}}(t) = \{ \tilde{\mathbf{p}}_{k,t}^* \mid k \in \mathcal{K}_t \},
\end{equation}
\ea{where $\mathcal{K}_t \subset \{ 1, \dots, \mathcal{C}\}$ denotes the subset of canonical markers successfully triangulated at time $t$. We apply different refinement strategies to address possible incompleteness due to self-occlusions, explained below:}

\newpage
\subsection{2D and 3D Linear Interpolation}
\label{sec:linear_interpolation}

\ea{To address possible short-term dropouts in marker tracking due to their small size, we apply linear interpolation in both the 2D and 3D domains. After performing 2D detection, gaps of length $\delta t \leq \tau_{\text{max}}$ in a camera stream $j$ are identified and filled using linear interpolation between the surrounding valid detections.  To preserve data plausibility, $\tau_{\text{max}}$ is kept low (i.e., 2 frames). The same strategy is applied to the 3D trajectories $\mathbf{p}_{k,t}^*$ obtained after triangulation. For any missing frame at time $t \in (t_1, t_2)$, the position is estimated as:}
\begin{equation}
\tilde{\mathbf{x}}_{t} = \frac{t_2 - t}{t_2 - t_1} \mathbf{x}_{t_1} + \frac{t - t_1}{t_2 - t_1} \mathbf{x}_{t_2}, \quad \text{for } t_1 < t < t_2   
\end{equation}
\ea{where $\mathbf{x}$ denotes either 2D image coordinates or 3D world coordinates.}

\subsection{Temporal Smoothing}
\label{sec:temporal_smoothing}

\ea{To suppress high-frequency flickering in marker trajectories, we apply a temporal median filter to the 3D corner positions. For each marker corner $k$, we consider its coordinate time series $\{\mathbf{p}_{k,t}^*\}_{t=1}^{T}$. We apply a median filter with an odd window size $w$, centered at each time step. The filtered trajectory $\tilde{\mathbf{p}}_{k,t}^*$ is computed component-wise as:} 
\begin{equation}
\tilde{\mathbf{p}}_{k,t}^* = \operatorname{median} \left( \left\{ \mathbf{p}_{k,t'}^* \;\middle|\; t' \in \left\{t - \frac{w-1}{2}, t + \frac{w-1}{2}\right\} \right\} \right)
\end{equation}

\subsection{Local Motion Consensus Filtering}
\label{sec:motion_consensus}

\ea{Although unlikely, spatial outliers may occur. Therefore, we apply a local motion consensus filter to evaluate the consistency of a marker's displacement relative to its spatial neighbors. For each marker $m$ at time $t$, we compute its centroid $\mathbf{c}_m^{(t)} \in \mathbb{R}^3$ as the mean of its four corner positions. Using a $k$-d tree constructed from the centroids in the previous frame, we identify the $K$ nearest neighbors of marker $m$ and compute their respective displacement vectors $\delta \mathbf{c}_n = \mathbf{c}_n^{(t)} - \mathbf{c}_n^{(t-1)}$.} 

\ea{The marker's own displacement $\delta \mathbf{c}_m$ is then compared against the median displacement of its neighborhood, $\mathbf{v}_{\text{med}} = \operatorname{median}(\{\delta \mathbf{c}_n\})$. A marker is classified as an outlier and excluded from the frame if the discrepancy $d_m$ exceeds a predefined threshold $\tau_{\text{out}}$:}
\begin{equation}
d_m = \| \delta \mathbf{c}_m - \mathbf{v}_{\text{med}} \| > \tau_{\text{out}}
\end{equation}
\section{Experimental Setup}


\subsection{Parameters Details}
Across all experiments, we adopt the following empirical parameters to train the SOMA framework and extract the resulting deformations.

\paragraph{Network Architecture} 
The core deformation networks representing the skin and muscle layers ($\mathcal{S}_{\text{bind}}$ and $\mathcal{M}_{\text{bind}}$) are implemented as U-Net-style encoder-decoder architectures with skip connections. Both networks consist of a two-layer encoder with hidden dimensions of $512$ and $1024$, a $1024$-dimensional bottleneck, and a two-layer decoder mirroring the encoder. To preserve the original pose signals and prevent vanishing gradients, the features from the encoder stages are concatenated to their corresponding decoder layers. We utilize GELU as the non-linear activation function across all hidden blocks, coupled with 1D Batch Normalization and Dropout ($p=0.1$). The input dimension, corresponding to the flattened relative rotation matrices of the 23 non-root joints, is 207, while the output layer linearly projects the decoder features to the $3N$ vertex displacements.

\paragraph{Training Hyperparameters}
The models are trained end-to-end using the Adam optimizer. We train for a total of 20 epochs with a batch size of 128 frames per iteration. The learning rate is held constant at $1 \times 10^{-4}$. Gradient clipping with a maximum norm of 1.0 is applied to prevent gradient explosion during optimization.

\paragraph{Loss Formulation and Regularization Weights}
To guarantee anatomically plausible deformations, continuous surface topology, and to prevent interpenetration between the individual muscles, our total loss function is governed by several weighted terms. The empirical weights for the full model are set as follows:
\begin{itemize}
    \item \textbf{Data/Reconstruction Loss ($\lambda_{\text{data}}$):} Set to $50$, this weight drives the primary vertex-to-vertex alignment between our network predictions and the ground-truth simulations in the canonical space.
    \item \textbf{Spatial Smoothness ($\lambda_{\text{smooth}}$ and $\lambda_{\text{biharmonic}}$):} We penalize high-frequency spatial gradients and unnatural bending using first-order Laplacian and second-order biharmonic constraints. For the muscle layer, both weights are set to $100$. To enforce the skin layer to act as a cohesive, thick tissue, its smoothness weights are set higher at $200$.
    \item \textbf{Stretch / Edge Length Preservation ($\lambda_{\text{spring}}$):} Set heavily to $1000$ for both the skin and muscle layers. This ensures the discretization of the meshes is maintained by penalizing deviations from the rest-pose edge lengths.
    \item \textbf{Tangential Sliding Penalty ($\lambda_{\text{tangent}}$):} To prevent vertices from freely sliding along the surface tangent, we apply a weight of $1.0$ for the muscle layer and $0.1$ for the skin layer.
    \item \textbf{Volume Preservation ($\lambda_{\text{vol}}$):} Set to $1000$ for both layers, aggressively mitigating unrealistic scaling and maintaining the geometric volume of the individual overlapping muscle prisms during severe articulations.
\end{itemize}

\subsection{Metrics}

We evaluate surface accuracy and internal anatomical plausibility using the following metrics:
\begin{itemize}
    \item \textbf{Global Surface Tracking:} Mean (MPME), Median (MedPME), and 90th percentile (P90) Euclidean errors between predicted ($\mathbf{p}_k^{\text{pred}}(\boldsymbol{\theta})$) and ground-truth ($\mathbf{p}_{k,t}^*$) markers. P90 specifically captures extreme deformations.
    \item \textbf{Dynamic Region Error (DRE):} To isolate complex soft-tissue dynamics from rigid regions, we compute tracking error exclusively on highly dynamic markers where baseline LBS error exceeds $\tau \in \{15, 20, 25\}$ mm.
    \item \textbf{Intersection Ratio (\%):} The percentage of inner vertices penetrating the outer layer, indicating anatomical collision avoidance.
    \item \textbf{Volume Preservation (\%):} The mean absolute percentage change from rest volume ($\frac{|V - V_0|}{V_0} \times 100$), measuring the isochoric plausibility of the generated biological tissues.
\end{itemize}

\subsection{Baselines}

We compare our method against:
\begin{itemize}
    \item \textbf{LBS:} Applied to the subject's static 3D scan $\mathcal{S}_{\text{bind}}$ and internal canonical muscle prediction $\mathcal{M}_{\text{bind}}$ without pose-dependent corrections.
    \item \textbf{HIT~\cite{kellerHITEstimatingInternal2024} (via SMPL~\cite{loperSMPLSkinnedMultiperson2015}):} Given fitted shape parameters and pose $\boldsymbol{\theta}$, HIT infers the SMPL skin and internal meshes. Comparing against this pipeline allows us to simultaneously evaluate the limitations of standard parametric surface blending (SMPL) and uncoupled volumetric prediction (HIT) against our explicitly coupled, high-resolution physical formulation.
\end{itemize}
\section{Prism Volume}
\label{sec:prism_volume}
We use $E_{\text{vol}}$ to regularize soft volume preservation. Let $P$ be a prism consisting of a lower triangle $\left(\vec{x}_1, \vec{x}_2, \vec{x}_3\right)$ and an upper triangle $\left( \vec{x}_4, \vec{x}_5, \vec{x}_6 \right)$. 
Using the FEM shape functions as defined in, e.g., Landau et al. \cite{landau2020theory}, the interior of this prism is given as the set of all points that can be written as
\begin{equation}
    \vec{x} = \xi\left({\alpha}\vec{x}_1+{\beta}\vec{x}_2+{\gamma}\vec{x}_3\right) +\left(1-\xi \right)\left({\alpha}\vec{x}_4+{\beta}\vec{x}_5+{\gamma}\vec{x}_6\right)
\end{equation}
with the constraints that $0 \leq \alpha, \beta, \gamma,\xi \leq 1$ and $\alpha+\beta +\gamma =1$.
The signed volume of a prism can be obtained as the integral over the volume defined by the prism's shape function:
\begin{align}
\int_{\xi=0}^1 \int_{\beta=0}^1 \int_{\alpha=0}^{1-\beta}
          \det\!\left(\mathbf J(\alpha, \beta, \xi)\right)\,
          \mathrm{d}\alpha\,\mathrm{d}\beta\,\mathrm{d}\xi,
\end{align}
where $\mathbf{J}$ is the Jacobian of the interpolation function
\begin{equation}
\mathbf{J}(\alpha,\beta,\xi) =
\begin{bmatrix}
\frac{\partial \vec{x}}{\partial \alpha} &
\frac{\partial \vec{x}}{\partial \beta} &
\frac{\partial \vec{x}}{\partial \xi},
\end{bmatrix}
\end{equation}
and the partial derivatives are given by
\begin{align}
  \frac{\partial \vec{x}}{\partial \alpha}&= \xi(\vec{x}_1 - \vec{x}_3) + (1 - \xi)(\vec{x}_4 - \vec{x}_6),\\
  \frac{\partial \vec{x}}{\partial \beta} &= \xi(\vec{x}_2 - \vec{x}_3) + (1 - \xi)(\vec{x}_5 - \vec{x}_6),\\
  \frac{\partial \vec{x}}{\partial \xi}   &= \alpha(\vec{x}_1 - \vec{x}_4) + \beta(\vec{x}_2 - \vec{x}_5) + (1-\alpha - \beta)(\vec{x}_3 - \vec{x}_6).
\end{align}
The determinant of this matrix is quadratic in $\xi$, which occurs in two columns, and linear in $\alpha$ and $\beta$, which occur in only one column. 

The integral can therefore be evaluated exactly using Gauss Legendre quadrature with two nodes for the integral over $\xi$ and one node for $\alpha$ and $\beta$, whose domain forms a triangle. The integral thus evaluates to 
\begin{equation}
  V = \tfrac{1}{4}\left(
    \det\!\left(\mathbf J(\vec{\omega}_1)\right) +
    \det\!\left(\mathbf J(\vec{\omega}_2)\right)
  \right),
\end{equation}
with nodes:
\begin{equation}
\begin{aligned}
\vec{\omega}_1 &= \left(\tfrac{1}{3}, \tfrac{1}{3}, \tfrac{1 - \sqrt{\tfrac{1}{3}}}{2}\right)^\top
\end{aligned}
\qquad\qquad
\begin{aligned}
\vec{\omega}_2 &= \left(\tfrac{1}{3}, \tfrac{1}{3}, \tfrac{1 + \sqrt{\tfrac{1}{3}}}{2}\right)^\top
\end{aligned}
\end{equation}

\section{Marker Visibility}

\begin{figure}[h]
\centering

\scriptsize
\setlength{\tabcolsep}{3pt}
\renewcommand{\arraystretch}{0.8}
\begin{tabular}{lccccccc}
\toprule
& S1--S5 & Tric.Ext. & Mil.Press & Curls & Deadlifts & Squats & Push‑ups \\
\midrule
\textbf{Cov. (\%)}  
& 82.1 & 94.6 & 91.9 & 85.2 & 74.6 & 73.9 & 65.0 \\
\bottomrule
\end{tabular}

\includegraphics[width=0.9\linewidth]{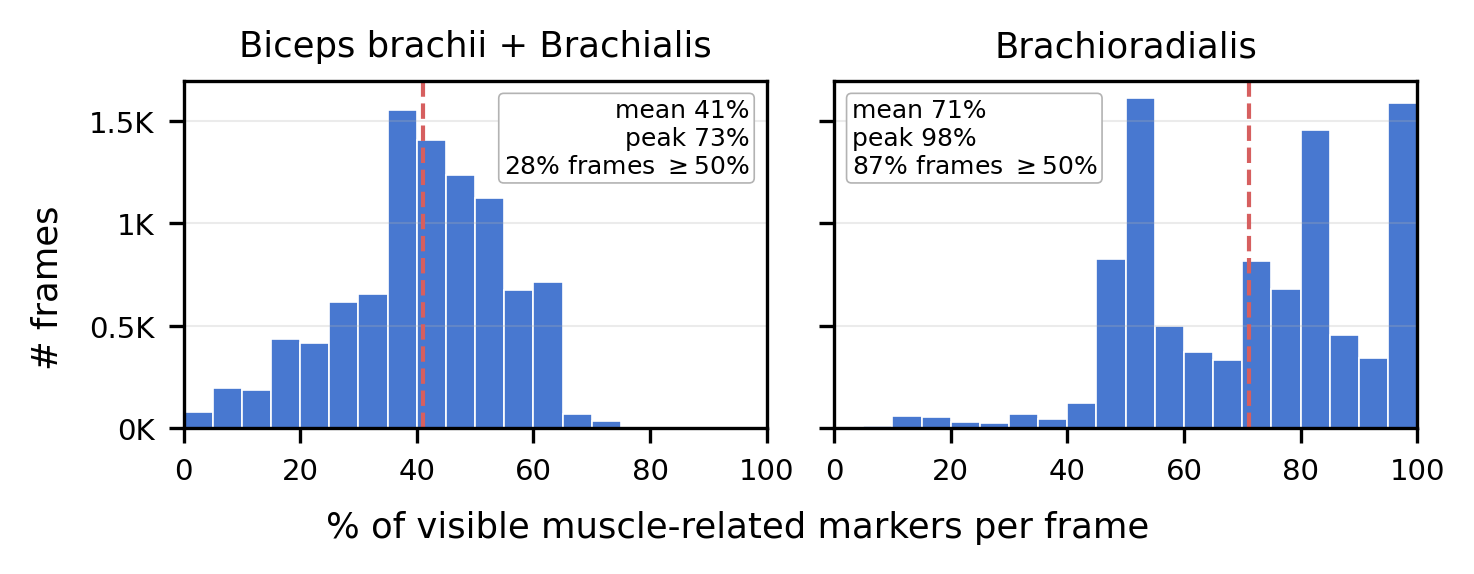}
\caption{Marker coverage and elbow flexors visibility for curls.}
\label{fig:coverage+visibility}
\end{figure}

Fig.~\ref{fig:coverage+visibility} reports mean marker detection coverage across all frames and 5~subjects.
We show a per-motion breakdown: Taking as an example arm-focused exercises, we retain 85--95\% coverage.
We complement this with a histogram of arm-related markers detected during curls.
The temporary reduction in \textit{biceps} and \textit{brachialis} visibility at peak flexion is a geometric consequence of the pose, yet lateral bulging remains visible, and the \textit{brachioradialis} (a synergistic flexor) stays well‑tracked. 

\section{Additional Qualitative Results}
\label{sec:qualitative_results}

\begin{figure*}
 \centering
 \includegraphics[width=1\linewidth,
                 trim=10 0 0 0, clip]{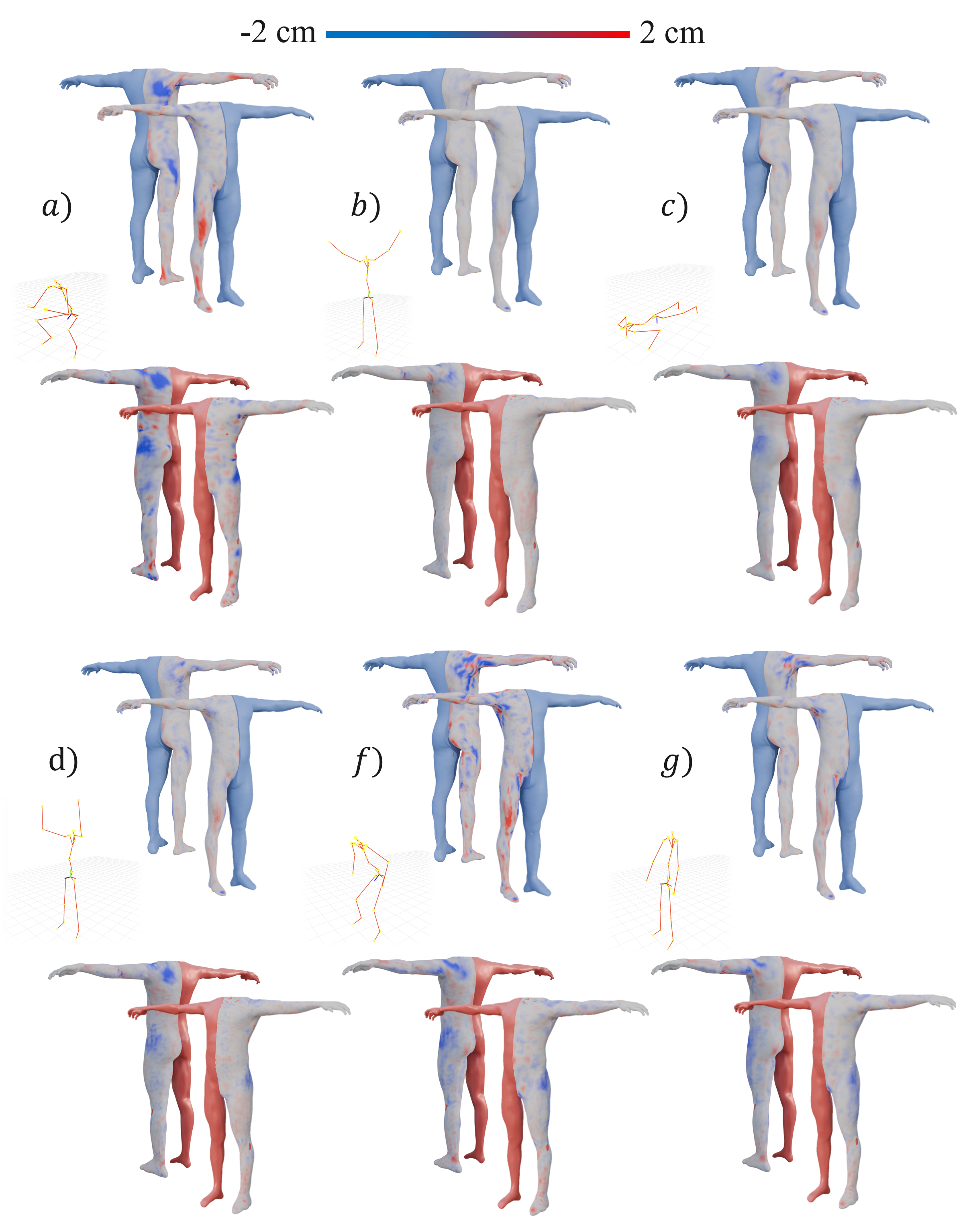}
\caption{\textbf{Predicted muscle and skin meshes across various activities.} (a) Squats, (b) Arm rotation, (c) Push-ups, (d) Military press, (e) Barbell bent-over rows, and (f) Deadlifts. For each activity, we visualize the displacement of both $\mathcal{S}_{\text{bind}}$ (top) and $\mathcal{M}_{\text{bind}}$ (bottom) from front and back viewpoints.}
 \label{fig:motions}
\end{figure*}

In Fig.~\ref{fig:motions}, we show different motions from our dataset, as well as the predicted inward (blue)/outward (red) deformations for both skin and muscle layers, $\mathcal{S}_{\text{bind}}$ and $\mathcal{M}_{\text{bind}}$ respectively.
We can observe how the corresponding muscles activate based on the exercise, s.t. quadriceps in the upper-leg during squats, the triceps during push-ups, or the shoulders for military presses. 


\begin{wrapfigure}[10]{r}{0.5\linewidth}
    \centering
    \vspace{-10mm}  \includegraphics[width=0.9\linewidth]{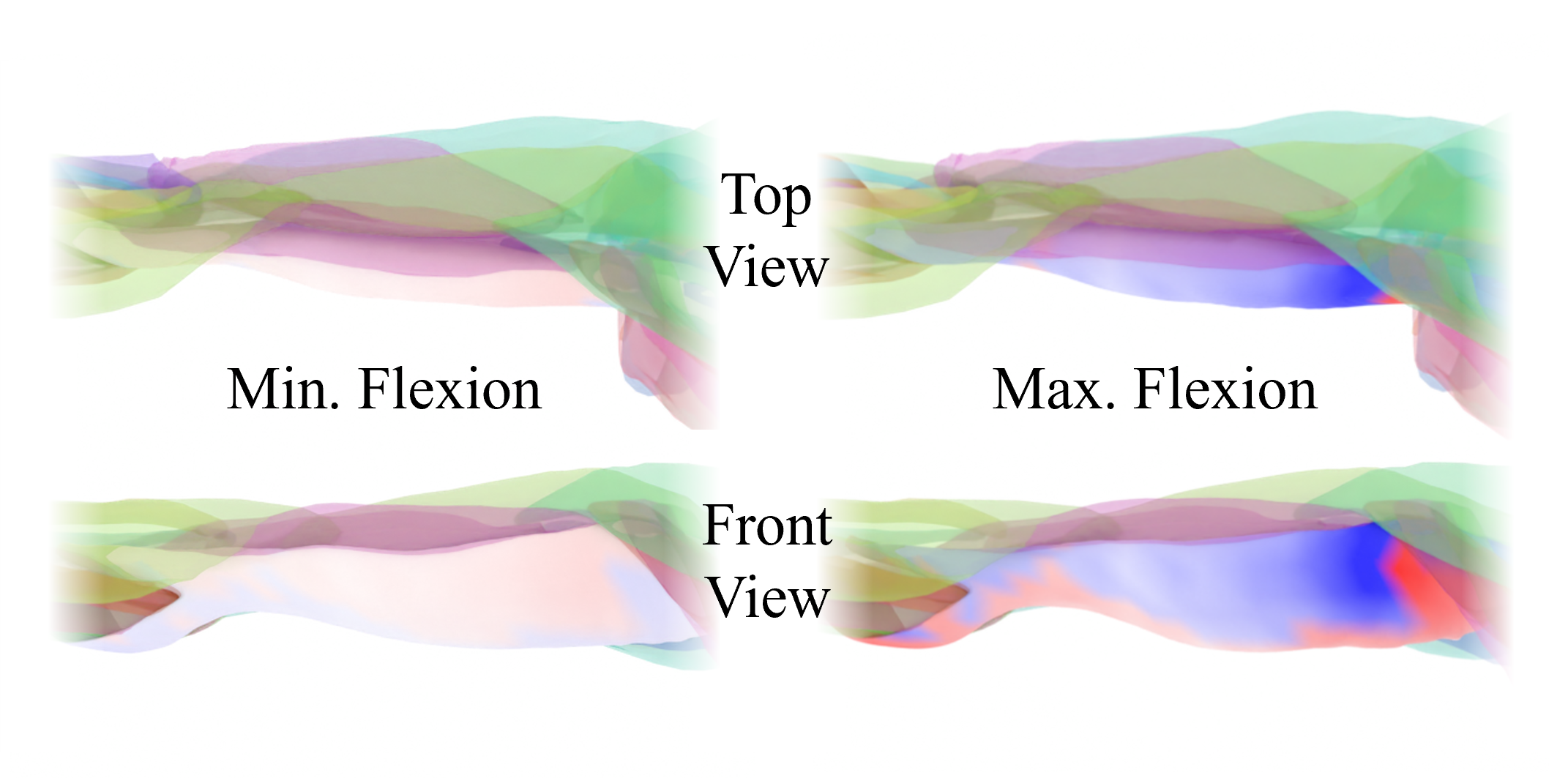}
SM/    \caption{\textbf{Biceps.} Deformation of biceps brachii during maximum and minimun flexion from top and front views.}
    \label{fig:bicep}
\end{wrapfigure}

As an example of our method on a specific muscle, we show the T-pose canonical bicep deformation in Fig.~\ref{fig:bicep}, using the frames of minimum and maximum forearm–upper-arm angle during curls. Following the original convention for positive (blue) and negative (red) deformation, the belly bulges outward while concavities form near the elbow and shoulder.

\section{Individual Muscle Deformation}

To animate detailed individual anatomical structures $\{\hat{\mathcal{M}}_{pred}^{(m)}\}_{m=1}^{M}$, we propagate the learned continuous deformation from the unified boundary layer $\mathcal{M}_{\text{bind}}$ to the underlying high-resolution meshes.

\ea{In a pre-processing step, we compute a binding that anchors every vertex of the high-resolution individual muscles to the canonical boundary muscle mesh $\mathcal{M}_{\text{bind}}$. For each vertex $\mathbf{u} \in \mathcal{M}_{\text{bind}}^{(m)}$ in the rest pose, we identify the closest triangle on $\mathcal{M}_{\text{bind}}$. Let $\mathcal{I}_u = \{k_1, k_2, k_3\}$ denote the indices of the three vertices forming this triangle, and $\mathbf{b}_u = \{b_{1}, b_{2}, b_{3}\}$ be the corresponding barycentric weights such that the projection of $\mathbf{u}$ onto the surface is $\mathbf{u}_{\text{proj}} = \sum_{j=1}^3 b_{j} \mathbf{v}_{k_j}$.}
During inference, we reconstruct the posed individual muscles by transferring the barycentric-interpolated canonical displacement to each internal vertex:
\begin{equation}
\tilde{\mathbf{u}}(\boldsymbol{\theta}) = \mathbf{u} + \sum_{j=1}^{3} b_{j} \mathbf{D}_{\text{musc}}(\mathbf{v}_{k_j}, \boldsymbol{\theta}).
\end{equation}
This formulation ensures that the internal high-resolution muscles deform in physical synchrony with the volumetric constraints enforced on the overall boundary structure of $\mathcal{M}_{\text{bind}}$.

\clearpage
%
%
\bibliographystyle{splncs04}
\bibliography{biblio}